\documentclass[10pt,twocolumn,letterpaper]{article}

\usepackage[pagenumbers]{cvpr} 

\usepackage{float}
\usepackage{pifont}
\usepackage{footnote}
\usepackage{enumitem}
\usepackage{bm}
\usepackage{arydshln}
\usepackage{booktabs}
\usepackage{multicol}
\usepackage{multirow}
\usepackage{color}
\usepackage{xcolor}     
\usepackage{colortbl}
\usepackage{soul}
\usepackage{bbding}
\usepackage{makecell}
\usepackage{mathtools}
\usepackage{imakeidx}
\usepackage{setspace}
\usepackage{threeparttable}
\makeindex
\usepackage{arydshln}
\usepackage{lipsum}
\usepackage{natbib}
\usepackage[toc]{multitoc}
\usepackage[edges]{forest}
\usepackage[normalem]{ulem}

\usepackage{bbding}
\usepackage[most]{tcolorbox}

\usepackage{algorithm}
\usepackage{algorithmic}

\usepackage{minitoc}
\usepackage[toc,page,header]{appendix}

\usepackage[T1]{fontenc}

\definecolor{orchid}{rgb}{0.85, 0.44, 0.84}
\definecolor{rubinred}{rgb}{0.82, 0.0, 0.28}
\definecolor{flagship}{rgb}{0.93, 0.06, 0.41}
\definecolor{radiologist}{rgb}{0.50, 0.50, 1}

\definecolor{DeepGreen}{RGB}{48,173,59}
\definecolor{LRed}{RGB}{230,88,34}

\newcommand{\method}{{\fontfamily{ppl}\selectfont
SMILE}}

\newcommand{\dataset}{{\fontfamily{ppl}\selectfont
CTVerse}}

\newcommand{\numofct}{1814}
\newcommand{\numofpatient}{477}
\newcommand{\numofclass}{88}
\newcommand{\numoftrainpatient}{85}
\newcommand{\numoftrainct}{312}
\newcommand{\numofmask}{159,632}
\newcommand{\numofhospital}{112}

\newcolumntype{P}[1]{>{\centering\arraybackslash}p{#1}}
\newlength\savewidth

\usepackage{graphicx}
\usepackage{caption}
\definecolor{cvprblue}{rgb}{0.21,0.49,0.74}
\usepackage[pagebackref,breaklinks,colorlinks,allcolors=cvprblue]{hyperref}

\def\paperID{195} 
\def\confName{CVPR}
\def\confYear{2026}

\title{See More, Change Less: Anatomy-Aware Diffusion for Contrast Enhancement}

\author{
Junqi Liu\textsuperscript{1,2} \quad
Zejun Wu\textsuperscript{1,3} \quad
Pedro R. A. S. Bassi\textsuperscript{1,4,5} \quad
Xinze Zhou\textsuperscript{1} \quad
Wenxuan Li\textsuperscript{1} \\
Ibrahim E. Hamamci\textsuperscript{6,7} \quad
Sezgin Er\textsuperscript{8} \quad
Tianyu Lin\textsuperscript{1} \quad
Yi Luo\textsuperscript{1} \quad
Szymon Płotka\textsuperscript{9} \\
Bjoern Menze\textsuperscript{6,7} \quad
Daguang Xu\textsuperscript{10} \quad
Kai Ding\textsuperscript{11} \quad
Kang Wang\textsuperscript{12} \quad
Yang Yang\textsuperscript{12} \\
Yucheng Tang\textsuperscript{10} \quad
Alan L. Yuille\textsuperscript{1} \quad
Zongwei Zhou\textsuperscript{1,}\thanks{Correspondence to Zongwei Zhou (\href{mailto:zzhou82@jh.edu}{\textsc{zzhou82@jh.edu}})} \\[2.5mm]
\textsuperscript{1}Johns Hopkins University \quad
\textsuperscript{2}University of Copenhagen \quad
\textsuperscript{3}University of Virginia \\
\textsuperscript{4}University of Bologna \quad
\textsuperscript{5}Italian Institute of Technology \quad
\textsuperscript{6}University of Zurich \\
\textsuperscript{7}ETH AI Center \quad
\textsuperscript{8}Istanbul Medipol University \quad
\textsuperscript{9}Jagiellonian University \\
\textsuperscript{10}NVIDIA \quad
\textsuperscript{11}Johns Hopkins Medicine \quad
\textsuperscript{12}University of California, San Francisco \\[1.5mm]
{\small Code, Dataset, and Models:~\href{https://github.com/MrGiovanni/SMILE}{https://github.com/MrGiovanni/SMILE}}
}

\begin{document}
\maketitle
\doparttoc 
\faketableofcontents 

\begin{abstract}

Image enhancement improves visual quality and helps reveal details that are hard to see in the original image. In medical imaging, it can support clinical decision-making, but current models often over-edit. This can distort organs, create false findings, and miss small tumors because these models do not understand anatomy or contrast dynamics. We propose \method, an anatomy-aware diffusion model that learns how organs are shaped and how they take up contrast. It enhances only clinically relevant regions while leaving all other areas unchanged. \method\ introduces three key ideas: (1) structure-aware supervision that follows true organ boundaries and contrast patterns; (2) registration-free learning that works directly with unaligned multi-phase CT scans; (3) unified inference that provides fast and consistent enhancement across all contrast phases. Across six external datasets, \method\ outperforms existing methods in image quality (\textbf{14.2\%} higher SSIM, \textbf{20.6\%} higher PSNR, \textbf{50\%} better FID) and in clinical usefulness by producing anatomically accurate and diagnostically meaningful images. \method\ also improves cancer detection from non-contrast CT, raising the F1 score by up to 10 percent.

\end{abstract}

\section{Introduction}\label{sec:intro}

\begin{figure}[t]
\centering
\includegraphics[width=\linewidth]{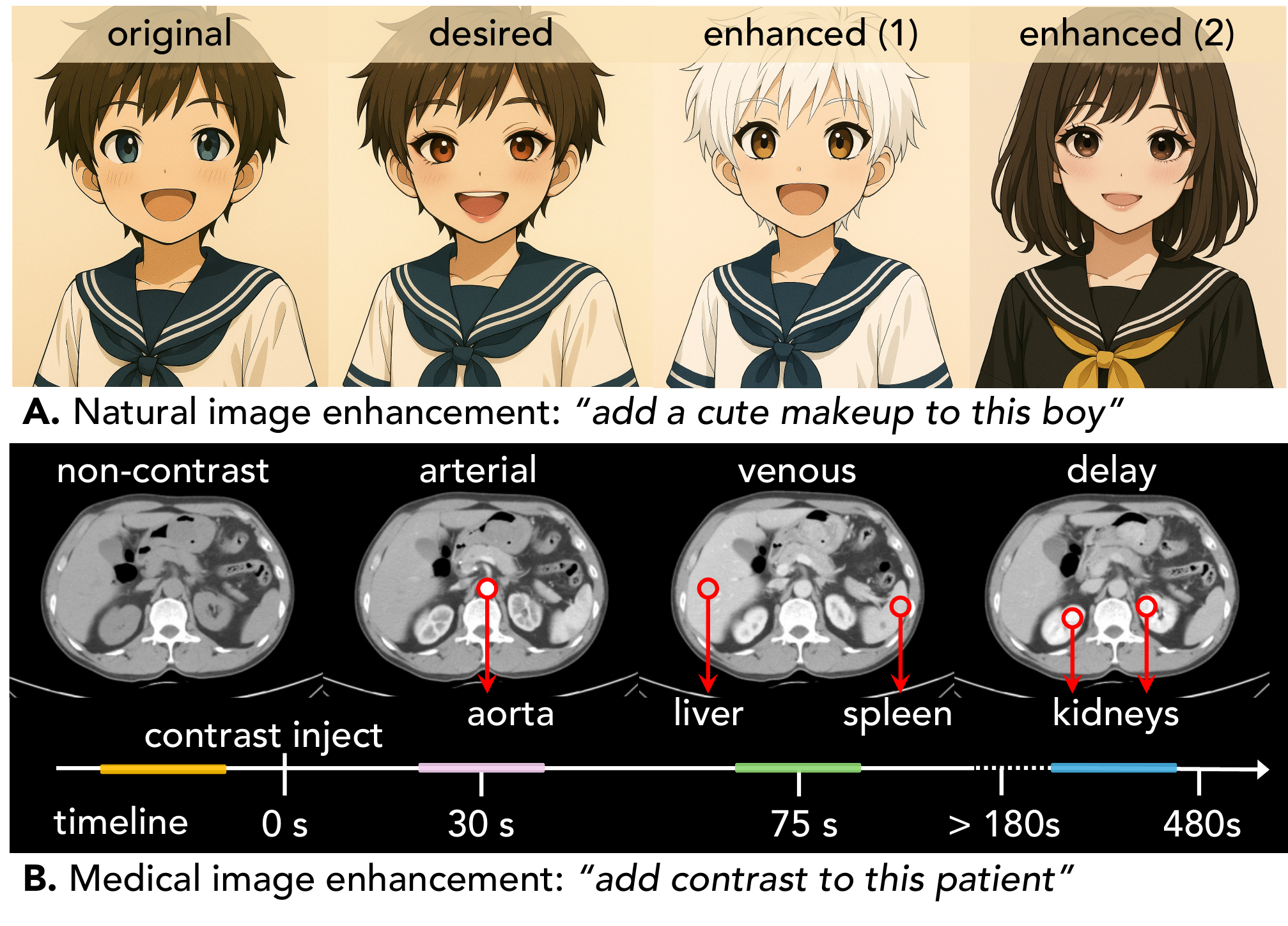}
    \caption{\textbf{A. Natural image enhancement.} When generative models add makeup to a photo, they often do too much---changing not just the face, but the hair and clothes as well. That’s fine for social media filters, but in medical imaging, such `over-creativity' can hide real tumors or create fake ones. \textbf{B. Medical image enhancement.} Doctors use contrast agents (a liquid injected into the body) to make internal organs easier to see. After injection, CT scans are taken at four times: (1) Non-contrast (\texttt{N}) is the baseline before enhancement; (2) Arterial (\texttt{A}) highlights arteries; (3) Venous (\texttt{V}) enhances organs such as the liver and spleen; and (4) Delay (\texttt{D}) shows mainly urinary system. The contrast makes certain tissues absorb more X-rays and appear brighter, revealing subtle tumors or vascular structures that would otherwise be invisible. These phases reflect how contrast flows through organs, helping radiologists detect and diagnose disease more accurately.
    }
\label{fig:teaser}
\end{figure}

Image enhancement is an important task in computer vision that aims to improve image quality or emphasize key details. Recent advances in image enhancement are fueled by generative AI, especially diffusion models \cite{zhang2023adding,ho2020denoising,li2024text,chen2024towards,chen2024analyzing,du2024boosting,chen2025scaling}, which can make enhancements through conditional generation. However, these models often become too creative, adding unwanted changes to the original image that reduce its usefulness. This happens because they lack clear knowledge of what objects are in the image and which parts should be changed. For example, think of how an AI model adds makeup to a face. Ideally, it should touch only what’s needed---the lips, skin tone, or eyes---while keeping everything else unchanged. Yet most generative models can’t resist being `creative': they also modify the hair, the clothes, or even the background \cite{siddiquee2019learning}, as shown in \figureautorefname~\ref{fig:teaser}A. These changes maybe okay in beauty filters, but dangerous in medicine. 
In CT scans, for example, a model that tries to enhance contrast\footnote{In medical imaging, contrast refers to a radiopaque substance injected into the bloodstream that temporarily increases tissue brightness, allowing radiologists to see blood vessels, organs, and potential tumors more clearly.} might unintentionally alter healthy organs or fabricate false tumors. Our goal is to build a model that knows exactly \textit{what to change} and \textit{what to keep}, ensuring every enhancement looks realistic---and remains clinically trustworthy. 

We hypothesize that if generative models can recognize objects and their fine structures (e.g., through semantic segmentation), they can focus edits on the right regions while keeping everything else unchanged.
This would enable image enhancement to make necessary, but minimal, changes. Clinically, this problem is highly relevant because approximately half of abdominal CT scans are acquired without contrast \cite{cao2023large,hu2025ai,bassi2025scaling}. If a reliable AI model could create realistic contrast-enhanced versions of these scans, it would make tumor detection easier and more accessible. 
Radiologists could identify tumors earlier without additional scans or injections. This highlights the need for enhancement models that are anatomically and physiologically accurate.

To study this, we first build a large-scale medical image dataset that offers two key advantages over existing datasets. \textbf{First,} our dataset includes per-pixel annotations of many organs, vessels, bones, and disease regions, enabling structural evaluation. In contrast, other datasets lack such detailed annotations of fine structures (e.g., hair, face, clothing), making them less suitable for assessing generative models. \textbf{Second,} our dataset provides paired scans for more than \numofpatient\ patients from over \numofhospital\ hospitals, each with non-contrast, arterial, venous, and delay phases (see \figureautorefname~\ref{fig:teaser}B). Other datasets, by comparison, lack paired “before-and-after” images. For example, in natural image datasets, there are no exact pairs of faces with and without makeup. We expect the insights from this study to generalize to natural image enhancement once comparable datasets become available for training and evaluation.

\begin{figure*}[ht]
    \centering
    \includegraphics[width=\linewidth]{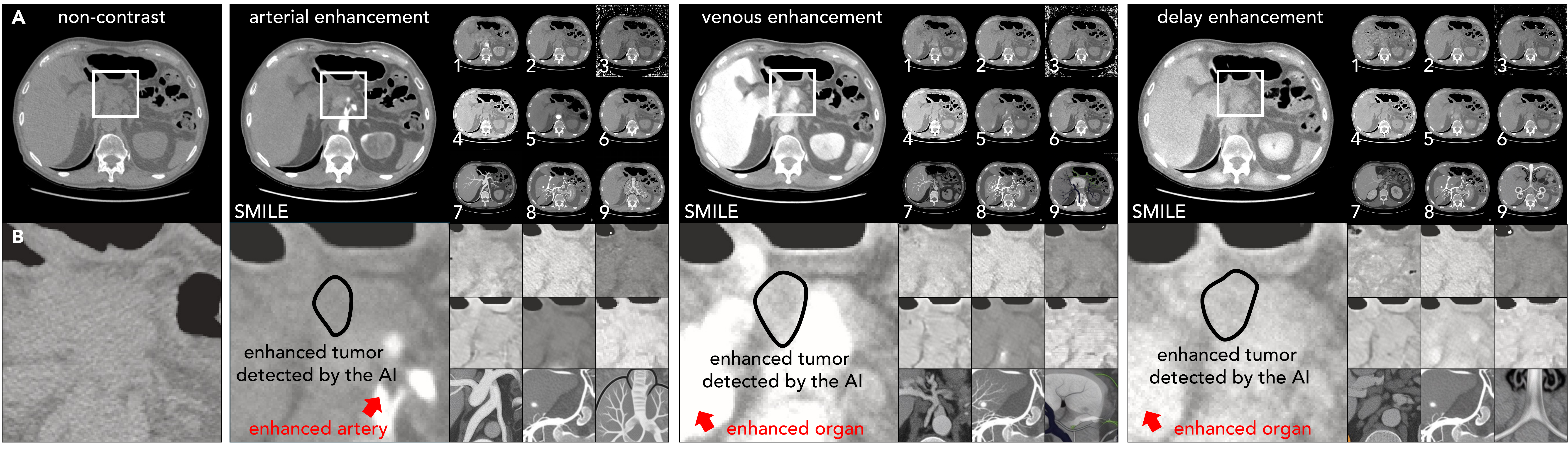}
    \caption{
    \textbf{\method\ ensures both structural consistency and intensity accuracy on enhanced scans, outperforming all competing models in both anatomical integrity and tumor visibility.}
    We evaluate whether the enhanced CT scans remain anatomically correct (no extra structures, no distortions, high image quality), and whether the added contrast is diagnostically correct by allowing the tumor to be clearly detected. 
    Baselines used (labeled 1–9):
    (1) Pix2Pix~\cite{isola2017image},  
    (2) CycleGAN~\cite{chu2017cyclegan},
    (3) CyTran~\cite{ristea2023cytran},  
    (4) DALL-E~\cite{esser2021taming},  
    (5) MedDiffusion~\cite{khader2023denoising},  
    (6) CUT~\cite{park2020contrastive},  
    (7) ChatGPT-5.1~\cite{chatgpt51},  
    (8) Google Nano Banana~\cite{google_nanobanana},  
    (9) Qwen-3 Max~\cite{qwen3max}.
    The last three large vision models are included only for visual reference to show that general image-editing systems cannot handle medical CT enhancement reliably.
    \textbf{A. Structural Consistency. }We evaluate \method's structural fidelity by comparing non-contrast CT enhancement results against 9 compelling baseline models. Compared to other methods, \method\ preserves organ boundaries and global anatomy without introducing extra structures or phase-inconsistent artifacts. This shows that \method\ maintains high structural reliability even under large intensity shifts. \textbf{B. Intensity Accuracy. }To further validate enhancement correctness, we apply a state-of-the-art tumor detector~\cite{li2025scalemai} on the enhanced arterial, venous, and delay phases. In all three enhanced phases, the tumor is successfully detected (marked by the black circle), confirming that \method\ restores clinically meaningful contrast cues needed for downstream diagnostic tasks. 
    }
    \label{fig:visualization}
\end{figure*}

We introduce \method\ (\ul{S}uper \ul{M}odality \ul{I}mage \ul{L}earning and \ul{E}nhancement), an anatomy-aware diffusion model designed for clinically reliable CT enhancement. Here, \emph{anatomy-aware} has two meanings. \textbf{First,} the enhanced image should still represent the same patient as the original. Many existing models change organ shapes or spatial structures so much that the result no longer looks like the same patient \cite{cai2024radiative,cai2024structure,mao2025medsegfactory} (see \figureautorefname~\ref{fig:visualization}A). \textbf{Second,} anatomy-aware means understanding basic physiology, i.e., how different organs, vessels, and tissues take up contrast over time. Tumors should neither disappear nor be artificially created. Current generative models completely lack this knowledge, often producing unrealistic contrast intensity that ignore the actual dynamics of the human body \cite{lin2025pixel,yang2025medical,guo2025text2ct,lai2024pixel,hu2022synthetic,hu2023label,li2023early,hu2023synthetic} (see \figureautorefname~\ref{fig:visualization}B).

\begin{figure}[t]
    \centering
    \includegraphics[width=\linewidth]{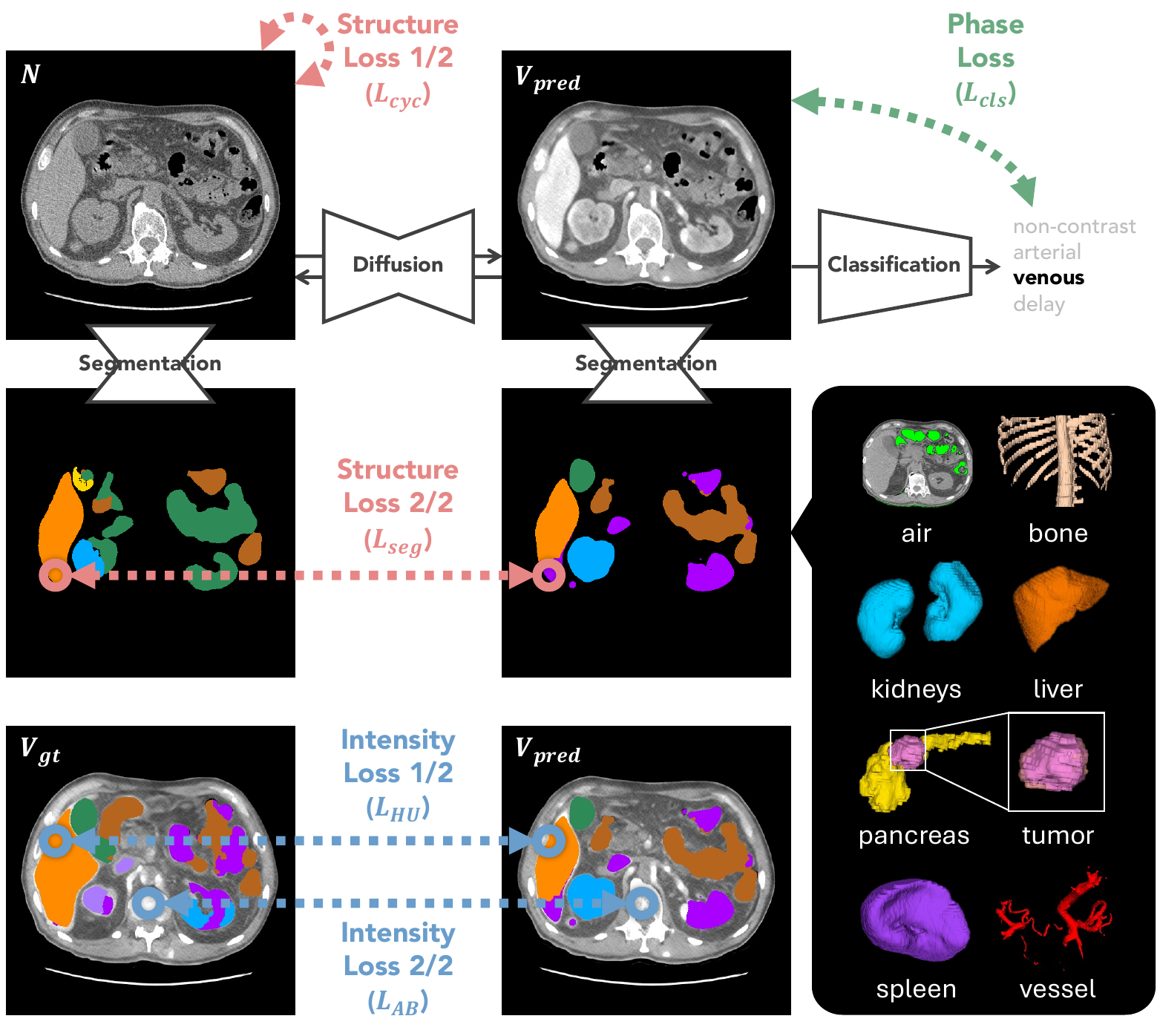}
    \caption{\textbf{\method\ employs anatomy-aware supervision.}
    Our framework integrates multiple anatomical constraints to guide contrast-phase enhancement. The structural segmentation loss ($\mathcal{L}_{\text{seg}}$) and cycle consistency loss ($\mathcal{L}_{\text{cyc}}$) preserve structural fidelity.
    The phase classification loss ($\mathcal{L}_{\text{cls}}$) ensures the enhanced CT shows the correct contrast-phase characteristics.
    Finally, the intensity HU loss $\mathcal{L}_{\text{HU}}$ and air/bone loss $\mathcal{L}_{\text{AB}}$ enforce realistic organ (mainly abdominal, e.g., liver) enhancement and maintain consistency in air and bone regions that should remain unchanged. 
    As demonstrated in the figure, \method\ does not require registration for enhancement source and ground truth.}
    \label{fig:method}
\end{figure}

\begin{figure*}[t]
    \centering
    \includegraphics[width=\linewidth]{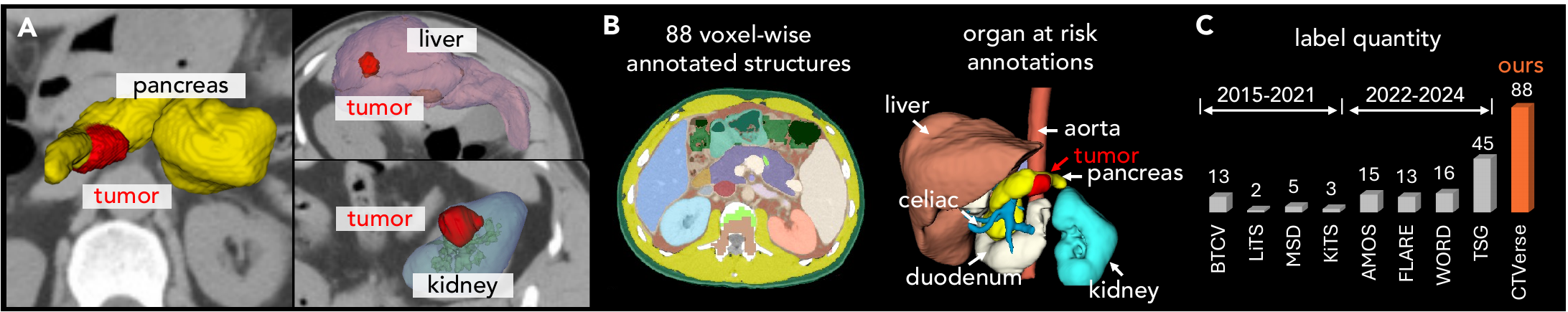}
    \caption{
    \textbf{\dataset\: sets a new standard for multi-phase CT benchmarks by offering the most comprehensive, finely annotated organ and tumor labels.} We introduce \dataset, a high-quality, multi-phase, and precisely annotated CT dataset designed for training and evaluating generative models. 
    \textbf{A.} \dataset\ provides detailed annotations for multiple tumor types (\textit{pancreas}, \textit{liver}, and \textit{kidney}). 
    \textbf{B.} \dataset\ contains voxel-wise annotations for 88 anatomical structures, including organs at risk, vessels, bones, etc. 
    \textbf{C.} Compared to existing publicly available datasets \cite{li2024abdomenatlas,li2025pants,bassi2025radgpt,qu2023annotating,li2024well,li2024medshapenet,chou2024embracing,liu2024universal,liu2023clip,chen2025vision}, our \dataset\ includes at least 1.5 times more labeled structures, making it a strong benchmark for generative models and medical research.
    }
    \label{fig:dataset_statistics}
\end{figure*}

To our knowledge, this is the \textit{first} study to quantitatively evaluate AI-generated CT enhancements for both \textbf{structural consistency} and \textbf{intensity accuracy}. Structural consistency ensures the enhanced image still represents the same patient, while intensity accuracy verifies that tissue density values remain clinically correct. Such evaluation was impossible before, as it requires precise organ and vessel masks—now enabled by our dataset and method.
In summary, we make two contributions that mark a significant step toward clinically useful contrast image enhancement:

\begin{enumerate}
    \item \textbf{An open, multi-phase dataset.} We present \dataset, a large CT dataset that includes four contrast phases (non-contrast, arterial, venous, and delay) for all \numofpatient\ patients from \numofhospital\ hospitals. Each scan is annotated with \numofclass\ anatomical structures and tumors in the pancreas, liver, and kidney, resulting in \numofmask\ three-dimensional masks (detailed statistics are in \figureautorefname~\ref{fig:dataset_statistics}). This dataset enables open-source training and quantitative evaluation of cross-phase enhancement methods.

    \item \textbf{An anatomy-aware method.} We develop \method, an anatomy-aware diffusion framework trained with multiple types of supervision to ensure both visual quality and clinical reliability (illustrated in \figureautorefname~\ref{fig:method}). The model introduces new loss functions that guide learning in three ways—preserving structure, maintaining clinical meaning, and ensuring realistic intensity. We conduct rigorously assessment not only in image quality but also in clinical relevance. Across six external datasets, \method\ achieves significant improvement in both image quality and clinical relevance: $+14\%$ SSIM, $+20\%$ PSNR, $+50\%$ FID, and $r>0.95$ in intensity correlation.

\end{enumerate}

We further demonstrate \method\ for opportunistic cancer screening using non-contrast CT scans---the most common type of abdominal imaging. These scans are widely available but often difficult for radiologists and AI systems to interpret because they lack contrast. Since many of these scans are taken for unrelated reasons, they represent a large and underused resource for early cancer detection. \method\ transforms non-contrast scans into realistic contrast versions (arterial, venous, or delay), revealing tumors that would otherwise remain invisible. When combined with existing detection models, the enhanced scans improve F1 score by 10\%, turning non-contrast CT scans into a practical and accessible tool for early cancer screening without additional cost or contrast exposure.

\section{\method}\label{sec:POC}

\method\ is a registration-free diffusion framework for multi-phase CT enhancement built on Stable Diffusion~\cite{rombach2022high}. It learns directly from unaligned data and uses three complementary supervision signals—structure, phase, and intensity—to guide the diffusion process. As shown in \figureautorefname~\ref{fig:method}, these anatomy-aware signals work together to preserve anatomy, enforce correct phase appearance, and maintain realistic tissue contrast. Overall, \method\ minimizes
\begin{equation}\footnotesize
\mathcal{L}_{\text{SMILE}}=\mathcal{L}_{\text{diff}} + \lambda_{\text{stu}}\mathcal{L}_{\text{stu}} + \lambda_{\text{cls}}\mathcal{L}_{\text{cls}} + \lambda_{\text{int}}\mathcal{L}_{\text{int}},
\end{equation}
where each term controls a specific form of structural or clinical supervision, enabling anatomically faithful and clinically reliable enhancement.

\subsection{Structural Loss}
Organ shapes vary noticeably across CT phases because the scans are not perfectly aligned. To handle this mismatch, \method\ adds structural supervision that does not require voxel-wise registration. The structural loss has two components: $\mathcal{L}_{\text{stu}} = \mathcal{L}_{\text{seg}} + \mathcal{L}_{\text{cyc}}$.

\smallskip\noindent\textbf{\textit{1.~Segmentation Loss:}} 
Because the shapes of anatomical structures such as organs, vessels, bones, etc., remain consistent across CT phases, \method\ adds a segmentation loss to preserve anatomy.
Both the enhanced CT $\hat{\mathbf{x}}_{\text{tgt}}$ and the input $\mathbf{x}_{\text{src}}$ are segmented using a pretrained nnU-Net~\cite{isensee2021nnu} on large, publicly available datasets \cite{li2024abdomenatlas,li2025pants,bassi2025radgpt,qu2023annotating,li2024well}.
The loss
$\mathcal{L}_{\text{seg}} = -\frac{1}{N}\sum G_i \log S_i(\hat{\mathbf{x}}_{\text{tgt}})$
penalizes structural differences between the enhanced and source masks.

\smallskip\noindent\textbf{\textit{2.~Cycle Consistency Loss:}}
Although CT phases differ in contrast appearance, their underlying anatomy remains the same.  
To reflect this property under unregistered training, \method\ applies a cycle reconstruction step in which the source phase $\mathbf{x}_{\text{src}}$ is recovered from the generated target phase $\hat{\mathbf{x}}_{\text{tgt}}$. 
Let $\boldsymbol{\epsilon}$ denote the sampled noise and $\boldsymbol{\epsilon}_{\theta}$ the corresponding prediction.  
The cycle loss is defined as
\begin{equation}\footnotesize
    \mathcal{L}_{\text{cyc}}
    = \|\hat{\mathbf{x}}_{\text{cyc}} - \mathbf{x}_{\text{src}}\|_1
    + \lambda_{\text{cdiff}}\|\boldsymbol{\epsilon} - \boldsymbol{\epsilon}_{\theta}\|_1,
\end{equation}
where the second term stabilizes diffusion-based reconstruction by enforcing noise-prediction consistency, and is computed only within the cycle path to regularize cycle diffusion process, separate from the main diffusion loss.

\subsection{Phase Loss}
Different phases differ mostly in local contrast changes (e.g., arterial enhancement in vessels or liver), while most regions of the CT are unchanged across phases. 
As a result, patch-based discriminators---which judge realism on small image crops---often receive little or ambiguous supervision, because most small crops look identical in all phases.
To address this, we simply add a pretrained phase classifier $C(\cdot)$ to guide the diffusion model, ensuring the generated image $\hat{\mathbf{x}}_{\text{tgt}}$ receives the correct phase label.

\subsection{Intensity Loss}
Contrast enhancement is mainly driven by intensity changes, so \method\ develops new losses that model how organ intensities should change or stay constant across phases, and is defined as $\mathcal{L}_{int}=\mathcal{L}_{HU}+\mathcal{L}_{AB}$.

\smallskip\noindent\textbf{\textit{1.~Organ HU Loss:}} 
Organ-level intensity patterns provide essential cues for distinguishing contrast phases.
Organ Hounsfield Unit (HU) values~\cite{denotter2019hounsfield,zhou2022interpreting} follow predictable shifts across contrast phases. \method\ computes a normalized squared error between the enhanced and target mean HU for each organ:
\begin{equation}\footnotesize
    \mathcal{L}_{\text{HU}} =
    \frac{1}{N_{\text{org}}}\sum_{i=1}^{N_{\text{org}}}
    \frac{(\bar{H}_{i}^{\text{pred}} - \bar{H}_{i}^{\text{gt}})^{2}}
    {(\bar{H}_{i}^{\text{gt}})^{2}},
\end{equation}
where $\bar{H}_{i}$ denotes the mean HU of the $i^{\text{th}}$ organ.  
This loss guides predicted organ intensities toward phase-consistent attenuation patterns.

\smallskip\noindent\textbf{\textit{2.~Air and Bone Loss:}} 
Air cavities and bones show minimal intensity variation across phases. A detector $D(\cdot)$ identifies these regions, and deviations between the enhanced and source images are penalized via:
\begin{equation}\footnotesize
\mathcal{L}_{\text{AB}}
= \| D(\hat{\mathbf{x}}_{\text{tgt}}) - D(\mathbf{x}_{\text{src}}) \|_2^2.
\end{equation}
This term constrains intensity stability in these regions while leaving contrast-sensitive structures free to vary.

\section{Experiments}\label{sec:exp}

\subsection{Datasets and Evaluation}
We test \method\ on a fully Out-of-Distribution (OOD) setup. The training and test data come from different sources, scanners, and patients. 
For evaluation, we use three common metrics: SSIM~\cite{wang2004image} to measure structural similarity, PSNR~\cite{korhonen2012peak} to measure image quality, FID~\cite{heusel2017gans} to measure how realistic the generated images look. 
Moreover, we are the first to use HU correlation to measure how well the organ intensity values match the real ones. 
Details of datasets used can be found in \tableautorefname\ref{tab:data_use}.

\begin{table}[t]
    \centering
    \scriptsize
    \caption{
    \textbf{Dataset characteristics.}
    Our constructed \dataset\ provides complete four-phase CT scans with full per-voxel annotations, whereas most public datasets contain only partial labels or limited phases.
    All public sources were checked for duplicates using 3D perceptual hashing.
    }
    \begin{tabular}{p{0.25\linewidth}P{0.12\linewidth}P{0.12\linewidth}P{0.13\linewidth}P{0.12\linewidth}}
    \toprule
    dataset     & patients & scans & phase & region   \\
    \midrule
    \rowcolor{flagship!10}
    \multicolumn{5}{l}{\emph{our constructed \dataset\ dataset, where \textbf{477} patients' scans will be released}}\\
    BRA~\cite{Kirk2016_TCGA-BLCA}     & 14        & 28    & \texttt{\textbf{A}} \texttt{\textbf{V}}           & BR         \\
    LiTS~\cite{BILIC2023102680}        & 18        & 36    & \texttt{\textbf{A}} \texttt{\textbf{V}}           & DE          \\
    WAW-TACE~\cite{doi:10.1148/ryai.240296}   & 163       & 652   & \texttt{\textbf{N}} \texttt{\textbf{A}} \texttt{\textbf{V}} \texttt{\textbf{D}}       & PL          \\
    MSD-CT~\cite{Antonelli_2022}      & 15        & 60    & \texttt{\textbf{N}} \texttt{\textbf{A}} \texttt{\textbf{V}} \texttt{\textbf{D}}       & US          \\
    VinDr~\cite{dao2022phase}      & 30        & 90   & \texttt{\textbf{N}} \texttt{\textbf{A}} \texttt{\textbf{V}}         & VN          \\
    PECN (private)    & 123       & 492   & \texttt{\textbf{N}} \texttt{\textbf{A}} \texttt{\textbf{V}} \texttt{\textbf{D}}       & CN          \\
    RTCN (private)     & 114       & 456   & \texttt{\textbf{N}} \texttt{\textbf{A}} \texttt{\textbf{V}} \texttt{\textbf{D}}       & CN          \\
    \rowcolor{flagship!10}
    \multicolumn{5}{l}{\emph{\textbf{85} patients used for AI training in this paper}}\\
    PACN (private)        & 24    & 96   & \texttt{\textbf{N}} \texttt{\textbf{A}} \texttt{\textbf{V}} \texttt{\textbf{D}}    & CN \\
    RTCN (private)     & 33    & 132   & \texttt{\textbf{N}} \texttt{\textbf{A}} \texttt{\textbf{V}} \texttt{\textbf{D}}    & CN \\
    FUS (private)        & 28    & 84    & \texttt{\textbf{N}} \texttt{\textbf{V}} \texttt{\textbf{D}}      & US \\
    \rowcolor{flagship!10}
    \multicolumn{5}{l}{\emph{\textbf{1,015} patients used for AI testing in this paper}}\\
    WAW-TACE~\cite{doi:10.1148/ryai.240296}   & 163   & 652   & \texttt{\textbf{N}} \texttt{\textbf{A}} \texttt{\textbf{V}} \texttt{\textbf{D}}       & PL  \\
    CT-RATE~\cite{hamamci2025developinggeneralistfoundationmodels}     & 74    & 74    & \texttt{\textbf{N}}             & TR  \\
    MSD-CT~\cite{Antonelli_2022}      & 15    & 60    & \texttt{\textbf{N}} \texttt{\textbf{A}} \texttt{\textbf{V}} \texttt{\textbf{D}}       & US \\
    PECN (private)     & 123   & 492   & \texttt{\textbf{N}} \texttt{\textbf{A}} \texttt{\textbf{V}} \texttt{\textbf{D}}       & CN  \\
    JUS (private) \cite{xia2022felix}        & 55    & 220   & \texttt{\textbf{N}} \texttt{\textbf{A}} \texttt{\textbf{V}} \texttt{\textbf{D}}       & US  \\
    TUS (private)        & 585   & 585   & \texttt{\textbf{N}}             & US  \\
    \bottomrule
    \end{tabular}
    \begin{tablenotes}
    \item BR: Brazil \quad CN: China \quad DE: Germany \quad PL: Poland \quad TR: Turkey
    \item US: United States \quad VN: Vietnam
    \end{tablenotes}
    \label{tab:data_use}
\end{table}

\begin{table*}[htbp]
    \centering
    \scriptsize
    \caption{
    \textbf{\method\ improves enhancement quality across all 12 phase conversions, with +14.2\% SSIM, +20.6\% PSNR, and 50.5\% better FID than the best baselines. In downstream tumor detection, \method\ further lifts F1-score by more than 10\%, while all baseline enhancement methods fail to provide any meaningful improvement.} 
    Quantitative comparison of multi-phase CT enhancement across 356 patients from 
    WAW-TACE~\cite{doi:10.1148/ryai.240296}, 
    MSD-CT~\cite{Antonelli_2022}, 
    PECN (private), 
    JUS (private) datasets.
    Each patient has four contrast phases: non-contrast (N), arterial (A), venous (V), and delay (D). For every source–target combination, representative scans were extracted from the CT volume after enhancement. We report SSIM, PSNR, and FID to assess structural similarity, perceptual fidelity, and distributional realism. The best and second-best results are marked with bold and underline. We also performed a one-sided Wilcoxon signed-rank test comparing our \method\ with all others, and statistically significant improvements at $P=0.05$ are highlighted in \textcolor{flagship!75}{pink}.
    }
    \begin{tabular}{p{0.09\linewidth}p{0.06\linewidth}P{0.034\linewidth}P{0.034\linewidth}P{0.034\linewidth}P{0.034\linewidth}P{0.034\linewidth}P{0.034\linewidth}P{0.034\linewidth}P{0.034\linewidth}P{0.034\linewidth}P{0.034\linewidth}P{0.034\linewidth}P{0.034\linewidth}P{0.08\linewidth}}
    \toprule
        & source & \multicolumn{3}{c}{\texttt{\textbf{N}}} & \multicolumn{3}{c}{\texttt{\textbf{A}}} & \multicolumn{3}{c}{\texttt{\textbf{V}}} & \multicolumn{3}{c}{\texttt{\textbf{D}}}\\
             \cmidrule(lr){3-5} \cmidrule(lr){6-8} \cmidrule(lr){9-11} \cmidrule(lr){12-14}
    method   & target & \texttt{\textbf{A}} & \texttt{\textbf{V}} & \texttt{\textbf{D}} & \texttt{\textbf{N}} & \texttt{\textbf{V}} & \texttt{\textbf{D}} & \texttt{\textbf{N}} & \texttt{\textbf{A}} & \texttt{\textbf{D}} & \texttt{\textbf{N}} & \texttt{\textbf{A}} & \texttt{\textbf{V}} & average \\
    \midrule
    \multirow{3}{*}{\makecell[l]{Pix2Pix~\cite{isola2017image}\\\textit{GAN-based}}} & SSIM ($\uparrow$) & 
    70.7 & 78.9 & 72.1 & 68.9 & 72.5 & 47.3 & 54.1 & 67.8 & 57.3 & 44.5 & 44.3 & 49.6 & 60.7\\
    & PSNR ($\uparrow$) & 
    18.9 & 21.3 & 19.3 & 19.9 & 21.0 & 14.9 & 18.4 & 20.7 & 19.9 & 15.9 & 16.5 & 19.0 & 18.8 \\
                 & FID  ($\downarrow$) & 
    147.8 & 179.5 & 377.3 & 423.7 & 232.2 & 250.9 & 414.3 & 324.2 & 322.0 & 416.6 & 228.9 & 279.4 & 299.7\\      
    \midrule
    \multirow{3}{*}{\makecell[l]{CycleGAN~\cite{chu2017cyclegan}\\\textit{GAN-based}}}    & SSIM ($\uparrow$)
    & 71.9 & 76.1 & 70.8 & 71.7 & 70.9 & 66.6 & 77.4 & 69.4 & 73.9 & 72.1 & 67.1 & 75.4 & 71.9\\
                 & PSNR ($\uparrow$) &
    17.9 & 19.3 & 17.2 & 18.3 & 18.0 & 16.4 & 20.8 & 17.9 & 18.8 & 18.4 & 16.6 & 19.3 & 18.2 \\
                 & FID  ($\downarrow$) &
    128.8 & 319.9 & 231.7 & 397.5 & 275.1 & 261.8 & 484.1 & 195.6 & 183.4 & 415.8 & 241.4 & 117.9 & 271.1 \\  
    \midrule
    \multirow{3}{*}{\makecell[l]{CyTran~\cite{ristea2023cytran}\\\textit{GAN-based}}}  
                & SSIM ($\uparrow$) &
    38.9 & 63.1 & 48.0 & 39.9 & 51.9 & 52.6 & 45.6 & 61.7 & 66.1 & 49.3 & 69.8 & 70.1 & 54.8 \\
                 & PSNR ($\uparrow$) &
    9.4 & 16.2 & 12.4 & 8.6 & 12.8 & 13.3 & 10.0 & 17.0 & 17.6 & 13.2 & 19.5 & 18.4 & 14.0 \\
                 & FID  ($\downarrow$) &
    193.5 & 231.2 & 258.8 & 373.4 & 300.3 & 292.5 & 374.0 & 178.3 & 244.0 & 407.8 & 209.7 & 254.4 & 276.5 \\ 
    \midrule
    \multirow{3}{*}{\makecell[l]{DALL\text{-}E~\cite{esser2021taming}\\\textit{VAE-based}}} 
                & SSIM ($\uparrow$) & 
    51.9 & 47.6 & 54.6 & 51.6 & 50.9 & 50.7 & 47.2 & 49.0 & 60.7 & 45.7 & 47.5 & 59.8 & 51.4\\
                & PSNR ($\uparrow$) &
    17.4 & 16.0 & 16.1 & 16.3 & 16.3 & 16.8 & 14.9 & 15.1 & 18.8 & 14.4 & 15.0 & 17.9 & 16.3 \\
                & FID  ($\downarrow$) &
    471.1 & 482.4 & 454.5 & 240.8 & 511.1 & 468.6 & 233.6 & 505.0 & 466.5 & 243.6 & 511.7 & 495.8 & 423.7\\      
    \midrule
    \multirow{3}{*}{\makecell[l]{MedDiffusion~\cite{khader2023denoising}\\\textit{DIFF-based}}}
                & SSIM ($\uparrow$) &
    63.2 & 72.2 & 72.0 & 69.1 & 62.4 & 62.7 & 55.3 & 60.3 & 67.2 & 64.3 & 70.4 & 55.7 & 64.6 \\
                & PSNR ($\uparrow$) &
    16.7 & 18.0 & 15.6 & 17.0  & 16.2 & 16.0 & 14.1 & 15.3 & 17.1 & 13.7 & 15.4 & 16.9 & 16.0\\
                & FID  ($\downarrow$) &
    490.0 & 298.9 & 299.4 & 468.5  & 516.1 & 503.1 & 310.9 & 506.5 & 205.8 & 298.6 & 482.4 & 495.3 & 406.3\\     
    \midrule
    \multirow{3}{*}{\makecell[l]{CUT~\cite{park2020contrastive}\\\textit{GAN-based}}} 
    & SSIM ($\uparrow$) & 66.8 & 73.6 & 70.6 & 72.9 & 77.3 & 79.1 & 78.7 & 73.9 & 80.1 & 75.5 & 75.3 & 81.3 & \ul{75.4}\\
    & PSNR ($\uparrow$) & 18.9 & 19.9 & 18.7 & 20.4 & 21.2 & 20.7 & 21.1 & 21.2 & 21.5 & 19.6 & 21.8 & 22.9 & \ul{21.4}\\
    & FID ($\downarrow$) & 283.4 & 258.6 & 280.9 & 363.8 & 204.4 & 259.3 & 333.4 & 182.3 & 270.7 & 360.1 & 172.5 & 265.1 & \ul{269.5}\\
    \midrule
    \multirow{3}{*}{\textbf{\method\ (ours)}} & SSIM ($\uparrow$) & \cellcolor{flagship!10}\textbf{85.1} & \cellcolor{flagship!10}\textbf{89.5} & \cellcolor{flagship!10}\textbf{82.9} & \cellcolor{flagship!10}\textbf{84.8} & \cellcolor{flagship!10}\textbf{86.9} & \cellcolor{flagship!10}\textbf{85.7} & \cellcolor{flagship!10}\textbf{86.0} & \cellcolor{flagship!10}\textbf{83.5} & \cellcolor{flagship!10}\textbf{86.1} & \cellcolor{flagship!10}\textbf{89.1} & \cellcolor{flagship!10}\textbf{85.6} & \cellcolor{flagship!10}\textbf{88.3} &
    \cellcolor{flagship!10}\textbf{86.1\tiny{~($\uparrow$14.2\%)}} \\
             & PSNR ($\uparrow$) & \cellcolor{flagship!10}\textbf{25.5} & \cellcolor{flagship!10}\textbf{27.5} & \cellcolor{flagship!10}\textbf{22.0} & \cellcolor{flagship!10}\textbf{24.4} & \cellcolor{flagship!10}\textbf{26.2} & \cellcolor{flagship!10}\textbf{25.0} & \cellcolor{flagship!10}\textbf{26.4} & \cellcolor{flagship!10}\textbf{25.5} & \cellcolor{flagship!10}\textbf{26.0} & \cellcolor{flagship!10}\textbf{28.1} & \cellcolor{flagship!10}\textbf{25.8} & \cellcolor{flagship!10}\textbf{27.3} &
             \cellcolor{flagship!10}\textbf{25.8\tiny{~($\uparrow$20.6\%)}} \\
             & FID  ($\downarrow$) & \cellcolor{flagship!10}\textbf{117.3} & \cellcolor{flagship!10}\textbf{152.8} & \cellcolor{flagship!10}\textbf{156.7} & \cellcolor{flagship!10}\textbf{148.6} & \cellcolor{flagship!10}\textbf{127.9} & \cellcolor{flagship!10}\textbf{129.2} & \cellcolor{flagship!10}\textbf{138.1} & \cellcolor{flagship!10}\textbf{134.3} & \cellcolor{flagship!10}\textbf{129.6} & \cellcolor{flagship!10}\textbf{112.6} & \cellcolor{flagship!10}\textbf{129.4} & \cellcolor{flagship!10}\textbf{124.8} &
             \cellcolor{flagship!10}\textbf{133.4\tiny{~($\downarrow$50.5\%)}}  \\
       
    \bottomrule
    \end{tabular}
    \begin{tablenotes}
    \item \texttt{\textbf{N}}: non-contrast \quad \texttt{\textbf{A}}: arterial \quad \texttt{\textbf{V}}: venous \quad \texttt{\textbf{D}}: delay \quad SSIM: structural similarity index measure \quad PSNR: peak signal-to-noise ratio \quad FID: Fr\'echet Inception Distance
    \end{tablenotes}
    \label{tab:enhancement}
\end{table*}

\smallskip\noindent\textbf{\textit{Training dataset:}}
We trained \method\ using a subset of the PACN (private), RTCN (private) and FUS (private) dataset, which provides multi-phase abdominal CT volumes acquired from the same patients under different contrast conditions. The training set consists of \numoftrainct\ CT volumes from \numoftrainpatient\ patients, organized as intra-patient multi-phase pairs to ensure physiological consistency across modalities.

\smallskip\noindent\textbf{\textit{Testing dataset:}}
Evaluation of \method\ is conducted on multiple independent external datasets, including WAW-TACE~\cite{doi:10.1148/ryai.240296}, CT-RATE~\cite{hamamci2025developinggeneralistfoundationmodels}, MSD-CT~\cite{Antonelli_2022}, PECN (private), JUS (private), and TUS (private).
For SSIM/PSNR comparison, all volumes are voxel-wise registered using ANTs~\cite{avants2011reproducible} solely to enable fair metric computation, since these metrics require phase-aligned images for valid comparison. The registration is not used during model training or inference, and does not affect \method’s generation process.

\smallskip\noindent\textbf{\textit{A new \dataset\ dataset (\figureautorefname~\ref{fig:dataset_statistics}):}} We will release a precisely annotated, phase-wise paired, and organ-wise registered high-quality dataset, \dataset, containing \numofct\ CT volumes from \numofpatient\ patients and voxel-wise annotations of 88 anatomical structures (details can be found in \tableautorefname~\ref{tab: CTVerse}). 
The data were collected from over \numofhospital\ hospitals, including 
BRA~\cite{Kirk2016_TCGA-BLCA},
LiTS~\cite{BILIC2023102680}, 
WAW-TACE~\cite{doi:10.1148/ryai.240296},
MSD-CT~\cite{Antonelli_2022},
VinDr~\cite{dao2022phase},
PECN (private),
and RTCN (private) datasets.
All volumes were registered at the voxel level using the ANTs~\cite{avants2011reproducible} toolkit to ensure accurate spatial alignment across phases. The precise voxel-level registration enables accurate comparison across contrast phases, providing a reliable benchmark for evaluating enhancement fidelity and structural consistency in image enhancement.

\subsection{Baseline and Implementation}
We compare \method\ with six representative baselines with the three main stream generative model architectures: 
(1) Generative Adversarial Network (GAN): Pix2Pix~\cite{isola2017image}, CycleGAN~\cite{chu2017cyclegan}, CUT~\cite{park2020contrastive} and CyTran~\cite{ristea2023cytran},
(2) Variational Autoencoder (VAE): DALL-E pytorch~\cite{esser2021taming},
(3) Diffusion Models (DIFF): MedDiffusion~\cite{khader2023denoising}. 
These methods were carefully selected because they represent the most widely adopted paradigms in image enhancement.

Other potentially relevant methods were not included for three justified reasons:
(1) \emph{Limited reproducibility:} Many recent works report results on private medical datasets without releasing code or pretrained models, making fair comparison impossible.
(2) \emph{Dataset mismatch:} Some methods are trained on  domain-specific data (e.g., brain MRI ), which do not generalize to CT enhancement task.
(3) \emph{Dimensional constraint:} Most image-enhancement methods are designed for 2D images and cannot handle 3D CT data with volumetric consistency required.

\method\ is trained for 100k steps on a single NVIDIA H100 (80 GB).
For implementation, \method\ uses five practical loss terms (\texttt{seg}, \texttt{cyc}, \texttt{cls}, \texttt{HU}, \texttt{AB}) although the method them into three fields (structure, phase, intensity). The distinction is explained in Appendix~\ref{sec: SMILE details}. We progressively enable supervision: diffusion only (0--2k), add phase+cycle (10k), add seg+HU+AB (20k). Fixed weights are
$\lambda_{\text{diff}}{=}1$, 
$\lambda_{\text{cls}}{=}10^{-3}$, 
$\lambda_{\text{cyc}}{=}10$, 
$\lambda_{\text{seg}}{=}10^{-3}$, 
$\lambda_{\text{HU}}{=}10^{-2}$, 
$\lambda_{\text{AB}}{=}1$, 
and become learnable after 80k via the Uncertainty Loss Module.

\section{Results}

\subsection{\textbf{\method} Exceeds Other Enhancement Methods}
We evaluate \method\ against six baselines on 
WAW-TACE~\cite{doi:10.1148/ryai.240296}, 
MSD-CT~\cite{Antonelli_2022}, 
PECN (private), 
and JUS (private) datasets.
Each case contains four phases (non-contrast, arterial, venous, delay). 
To keep evaluation efficient while still reflecting whole-volume quality, we use uniformly sampled scans per source–target pair, which reliably represent overall CT enhancement without unnecessary computational cost.
\tableautorefname~\ref{tab:enhancement} reports SSIM, PSNR, and FID across all 12 enhancement directions. 
\method\ achieves the highest scores on all metrics and all phase conversions. 

\subsection{\textbf{\method} Maintains Anatomical Structures}
\begin{figure*}[ht]
    \centering
    \includegraphics[width=\linewidth]{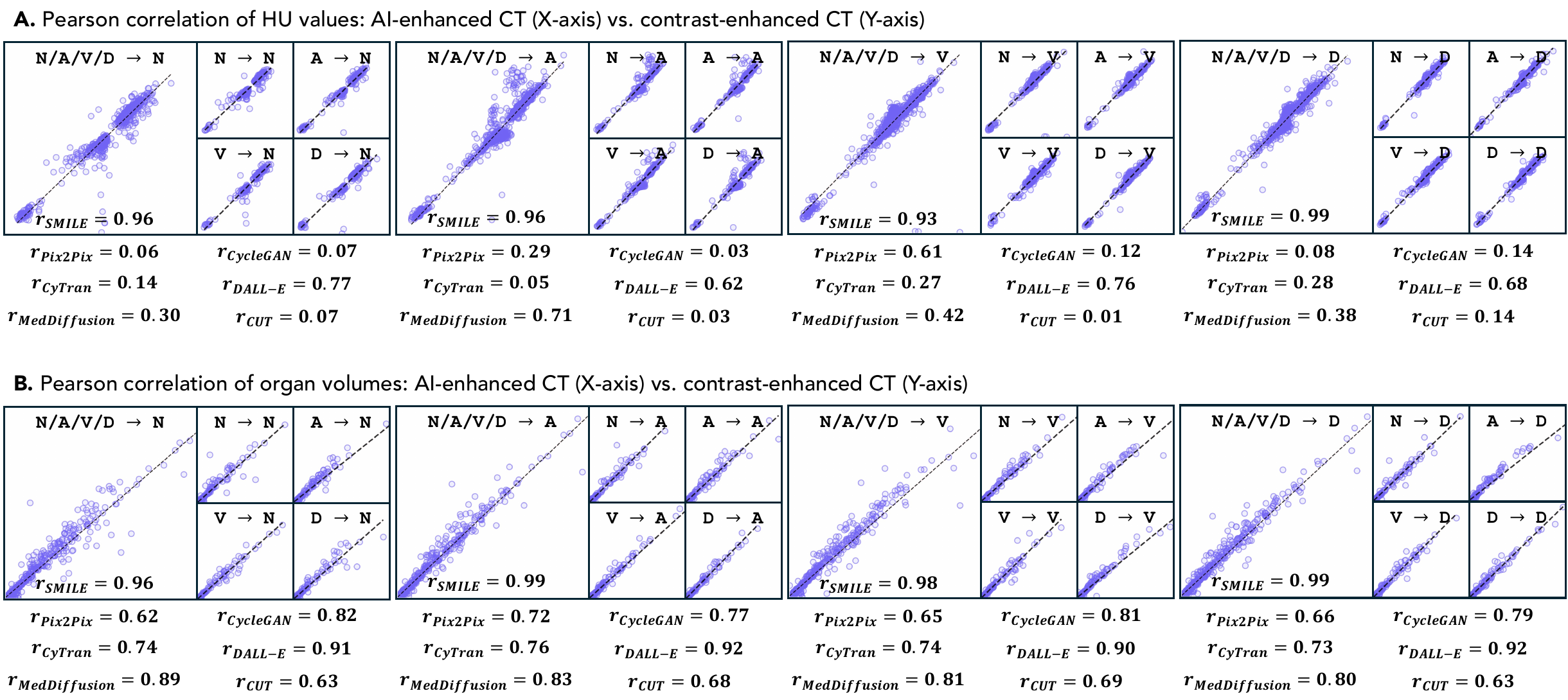}
    \caption{\textbf{Across all phases and organs, \method\ achieves an average HU and size correlation above 0.95, demonstrating strong anatomical and intensity consistency with real CT scans.} 
    We evaluate how well \method\ preserves organ intensity (HU) and volume size compared with the real CT scans across 22 organs.
    Here the HU is averaged over the whole organ rather than per-pixel.
    Each large plot corresponds to a target phase, N: non-contrast, A: arterial, V: venous, D: delay, and the four small plots on the right show results as enhanced from other source phases.
    The smaller plots illustrate all source→target phase pairs, showing that \method\ delivers consistently strong organ HU and size alignment regardless of the enhancement direction.
    Overall, \method\ maintains high consistency in both HU and organ size, demonstrating stable and reliable enhancement performance.
    }
    \label{fig:correlation}
\end{figure*}

We evaluate \method\ on 22 annotated organs\footnote{The full list is provided in Appendix \tableautorefname~\ref{tab:22organ}.} across all four contrast phases from
WAW-TACE~\cite{doi:10.1148/ryai.240296}, 
MSD-CT~\cite{Antonelli_2022}, 
PECN (private), 
and JUS (private) datasets.
\figureautorefname~\ref{fig:correlation} shows the correlation between synthesized and real CT scans in organ-level mean HU and volume. Mean HU is computed per organ, providing a stable and clinically meaningful metric. All scans are segmented using the state-of-the-art VISTA-3D~\cite{he2025vista3d} to obtain accurate organ measurements.
\method\ closely matches real anatomical statistics, achieving average correlations $>$0.95 for both HU and size. This demonstrates that \method\ enhances image quality while preserving organ integrity and clinical realism across all phases.

\begin{figure*}[ht]
    \centering
    \includegraphics[width=\linewidth]{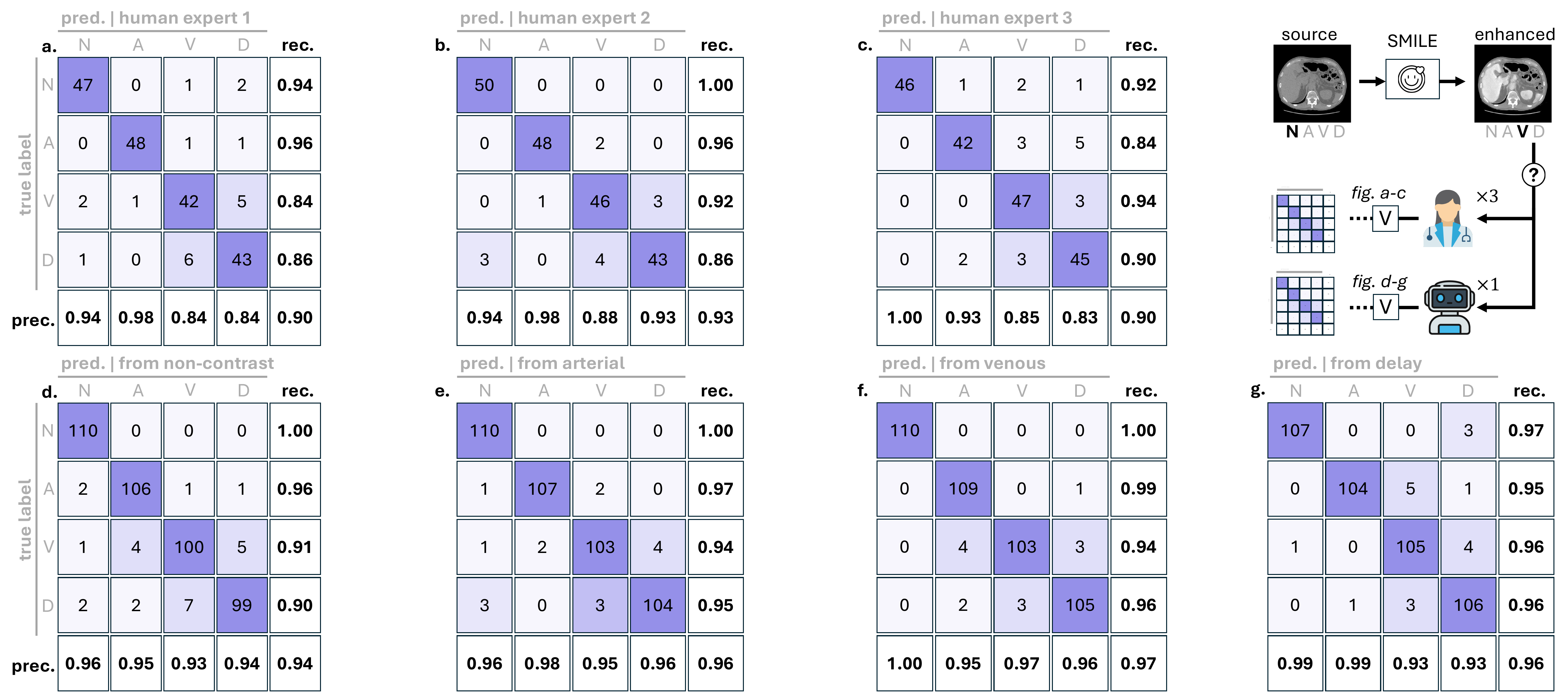}
    \caption{
    \textbf{\method\ consistently produces accurate and distinguishable phase-specific enhancements, achieving average over 95.0 precision and recall score.}
    \textbf{(a–c)} Reader study: three radiologists from two hospitals were shown a mixed set of \method-enhanced CT scans containing all four phases (N: non-contrast, A: arterial, V: venous, D: delayed) shuffled together. 
    Each radiologist independently labeled the phase of every slice, and we plot confusion matrices from their annotations.
    \textbf{(d–g)} Classification model:  We also evaluated \method\ phase enhancement correctness via a pretrained phase classifier. 
    By feeding \method-enhanced scans enhanced from each source phase (N from N/A/V/D, A from N/A/V/D, V from N/A/V/D, D from N/A/V/D), and plot the confusion metrics respectively. 
    }
    \label{fig:confusion_matrix}
\end{figure*}

\begin{figure}[htbp]
    \centering
    \includegraphics[width=\linewidth]{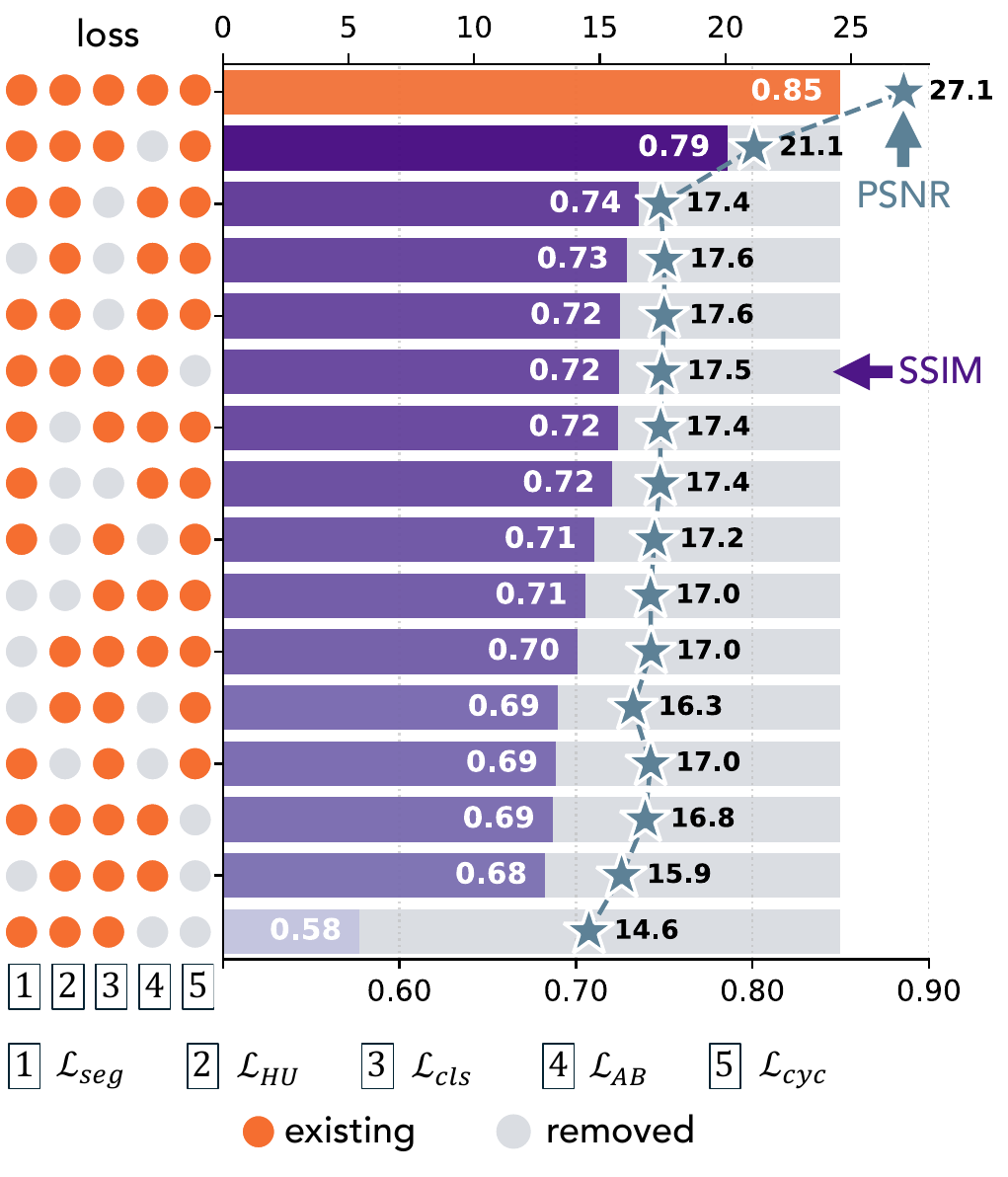}
    \caption{
    \textbf{Structural and intensity losses drive the largest gains.} Removing any loss module reduces performance, confirming that all five components contribute to \method's stability. The biggest drops occur when structural losses ($\mathcal{L}_{\text{seg}}$, $\mathcal{L}_{\text{cyc}}$) are removed, showing they are essential for preserving anatomy. Intensity losses ($\mathcal{L}_{\text{HU}}$, $\mathcal{L}_{\text{AB}}$) and the phase loss ($\mathcal{L}_{\text{cls}}$) also provide strong improvements by enforcing realistic contrast behavior, while phase supervision adds a smaller but consistent benefit.
    }
    \label{fig:ablation}
\end{figure}

\subsection{\textbf{\method} Enhances Precise Contrasts}
We assess phase accuracy using a radiologist reader study and a pretrained phase classifier (\figureautorefname~\ref{fig:confusion_matrix}). Both tests show that \method\ produces clear and correct phase-specific enhancements, with overall precision and recall above 95\%.

\smallskip
\noindent\textbf{\textit{Reader study.}}
Three radiologists from two hospitals labeled a mixed set of \method-enhanced scans containing all four phases. Their confusion matrices show that each phase is easily and consistently identified.

\smallskip
\noindent\textbf{\textit{Phase classification model.}}
A pretrained DenseNet~\cite{huang2017densely} classifier further evaluates phase correctness. For each source phase, we classify all \method-enhanced outputs, and the resulting confusion matrices show high precision and recall across all source–target phase pairs.

\begin{table}[t]
    \centering
    \scriptsize
    \caption{
    \textbf{\method\ improves early cancer detection with 10–20\% performance gains using non-contrast CT scans.}
    We evaluate \method\ as a preprocessing module for two state-of-the-art pancreatic tumor detectors, MedFormer~\cite{gao2022data} and ScaleMAI~\cite{li2025scalemai}, on external non-contrast CT scans.
    Adding \method\ yields strong gains in sensitivity, f1-score, and AUC with only modest drops in specificity, demonstrating that \method\ consistently improves the performance of these SOTA detectors.
    }
    \begin{tabular}{p{0.18\linewidth}P{0.045\linewidth}P{0.05\linewidth}P{0.045\linewidth}P{0.045\linewidth}P{0.045\linewidth}P{0.045\linewidth}P{0.045\linewidth}P{0.045\linewidth}}
    \toprule
     & \multicolumn{4}{c}{CT-RATE ($N=74$)} & \multicolumn{4}{c}{TUS ($N=585$)} \\
     \cmidrule(lr){2-5}\cmidrule(lr){6-9} 
    method   & Sen.  & Spec. & F1. & AUC  & Sen.   & Spec. & F1.  & AUC \\
    \midrule
    Gao~\etal~\cite{gao2022data}
    & 53.9 & 95.1 & 60.9 & 0.75 
    & 78.0 & 98.7 & 81.2 & 0.88\\
    \quad +\method 
    & 69.2 & 91.8 & 71.4 & 0.84 
    & 90.0 & 98.7 & 88.2 & 0.94 \\
    \quad $\mathbf{\Delta}$   
    & \textcolor{DeepGreen}{+15.3} 
    & \textcolor{LRed}{-3.3} 
    & \textcolor{DeepGreen}{+10.5}
    & \textcolor{DeepGreen}{+0.09}
    & \textcolor{DeepGreen}{+12.0}
    & 0.0
    & \textcolor{DeepGreen}{+7.0}
    & \textcolor{DeepGreen}{+.06}\\ 
    \midrule
    Li~\etal~\cite{li2025scalemai} 
    & 61.5 & 86.9 & 55.2 & 0.74 
    & 86.0 & 91.9 & 63.3 & 0.88      \\
    \quad +\method\ 
    & 84.6 & 80.3 & 61.1 & 0.83
    & 94.0 & 91.9 & 66.1 & 0.92      \\
    \quad $\mathbf{\Delta}$   
    & \textcolor{DeepGreen}{+23.1}
    & \textcolor{LRed}{-6.6}
    & \textcolor{DeepGreen}{+5.9}
    & \textcolor{DeepGreen}{+0.09}
    & \textcolor{DeepGreen}{+8.0}
    & 0.0
    & \textcolor{DeepGreen}{+2.8}
    & \textcolor{DeepGreen}{+.04}\\ 
    \bottomrule
    \end{tabular}
    \begin{tablenotes}
    \item Sen.: sensitivity \quad Spec.: specificity \quad F1.: f1-score
    \item AUC: area under the receiver operating characteristic curve
    \end{tablenotes}
    \label{tab:tumor}
\end{table}

\subsection{\textbf{\method} Detects Tumors in Non-Contrast Scans}

We evaluate \method\ as a preprocessing module for two pancreatic tumor detectors: MedFormer~\cite{gao2022data} (winner of the Touchstone benchmark \cite{bassi2024touchstone}) and ScaleMAI~\cite{li2025scalemai}. 
Both models represent the current state of the art in pancreatic tumor analysis, achieving strong detection and segmentation performance in recent benchmark studies. 
MedFormer reports 82–88\% F1 on multi-institutional benchmarks, while ScaleMAI achieves over 90\% sensitivity for early pancreatic lesions~\cite{gao2022data,li2025scalemai}.
We evaluate both detectors on external non-contrast CT scans from CT-RATE~\cite{hamamci2025developinggeneralistfoundationmodels} and TUS (private), and report sensitivity, specificity, F1, as well as AUC in \tableautorefname~\ref{tab:tumor}.
Adding \method\ increases sensitivity, f1-score and AUC across all settings, with modest changes in specificity, indicating improved early tumor detectability without introducing false-positive drift. These results suggest that \method\ can serve as a general enhancement module to improve a wide range of tumor detection systems operating on low-contrast CT scans.

\subsection{Ablation Study of \textbf{\method} Components}
We conduct an ablation study on 40 patients from our test datasets to assess the contribution of each loss module in \method. A small paired subset is sufficient to show the effect of each component. The model uses five losses: $\mathcal{L}_{\text{seg}}$ and $\mathcal{L}_{\text{cyc}}$ for structure, $\mathcal{L}_{\text{cls}}$ for phase, and $\mathcal{L}_{\text{HU}}$ and $\mathcal{L}_{\text{AB}}$ for intensity. In each experiment, we remove one or two losses and train the model under the same settings as \method. As shown in \figureautorefname~\ref{fig:ablation}, removing any module reduces SSIM and PSNR, demonstrating that all five losses are essential for stable, high-quality enhancement.

\section{Related Work}\label{sec:related_works}

For CT phase translation, existing image enhancement and translation methods (summarized in \tableautorefname~\ref{tab:enhancement}) do not meet three key needs at the same time:
(1) preserving anatomy,
(2) handling unregistered multi-phase CT scans, and
(3) producing clinically meaningful enhancement.

\smallskip\noindent\textbf{\textit{Natural image enhancement.}}
Paired methods, like Pix2Pix \cite{isola2017image}, rely on pixel-level losses and require perfectly aligned data, which is unrealistic for multi-phase CT because scans from different contrast phases are rarely voxel-aligned.
Unpaired models, like CycleGAN~\cite{zhu2017unpaired}, MUNIT~\cite{huang2018multimodal}, and CUT~\cite{park2020contrastive}, remove the need for alignment, but often cause geometric shifts or hallucinated structures.
Diffusion-based methods (DiffuseIT~\cite{kwon2022diffusion}, UNIT-DDPM~\cite{sasaki2021unit}) improve visual quality, yet their evaluation focuses on appearance rather than anatomical or clinical accuracy.

\smallskip\noindent\textbf{\textit{Medical image enhancement.}}
In medical imaging, unpaired translation models have been extensively studied for 
cross-modality and cross-contrast synthesis (e.g., MRI–CT or T1–T2)~\cite{chen2024contourdiff, ozbey2023unsupervised, wang2024structure, phan2024structural, luo2024target}.  
ContourDiff~\cite{chen2024contourdiff} introduces edge-aware constraints to better preserve structure, 
while SynDiff~\cite{ozbey2023unsupervised} incorporates diffusion-based priors to improve robustness under unpaired settings. However, these methods are not specialized for CT contrast-phase enhancement, where HU values impose physical constraints and phase-dependent changes are confined to a limited set of anatomically meaningful regions.
Recent GAN and diffusion models for CT contrast synthesis~\cite{zhu2025contrast, yin2025enhanced, xiao2022contrast, gardezi2024resnct, zhucausality, zhu2025cross} depend on voxel-level supervision or pre-registration, which is rarely available in routine clinical settings.
CyTran~\cite{ristea2023cytran} can operate without pairing, but still requires post-hoc registration.

\section{Conclusion and Discussion}

We introduced \method, an anatomy-aware diffusion framework that enhances CT images by modifying only clinically relevant regions and preserving all unchanged anatomy. With structural, phase, and intensity supervision, \method\ generates realistic contrast-enhanced images that follow true organ geometry and physiological patterns without creating false structures. The results are both visually convincing and clinically reliable.

\method\ also addresses an important clinical need. Many patients cannot receive contrast due to kidney issues, allergies, or unstable conditions, leaving clinicians to interpret limited non-contrast scans. \method\ provides a safe alternative by producing diagnostic-quality contrast images directly from non-contrast CT, without extra injections or radiation. This can improve care for vulnerable patients, aid rapid decisions in emergency settings, and simplify radiation oncology workflows by removing the need for cross-phase registration.

\newpage
\noindent\textbf{Acknowledgments.}
This work was supported by the Lustgarten Foundation for Pancreatic Cancer Research and the National Institutes of Health (NIH) under Award Number R01EB037669. We would like to thank the Johns Hopkins Research IT team in \href{https://researchit.jhu.edu/}{IT@JH} for their support and infrastructure resources where some of these analyses were conducted; especially \href{https://researchit.jhu.edu/research-hpc/}{DISCOVERY HPC}. We thank Jaimie Patterson for writing a \href{}{news article} about this project. Paper content is covered by patents pending.

{
    \small
    \bibliographystyle{ieeenat_fullname}
    \bibliography{refs,zzhou}

\begin{thebibliography}{76}
\providecommand{\natexlab}[1]{#1}
\providecommand{\url}[1]{\texttt{#1}}
\expandafter\ifx\csname urlstyle\endcsname\relax
  \providecommand{\doi}[1]{doi: #1}\else
  \providecommand{\doi}{doi: \begingroup \urlstyle{rm}\Url}\fi

\bibitem[{Alibaba Qwen Team}(2025)]{qwen3max}
{Alibaba Qwen Team}.
\newblock Qwen-3 max.
\newblock \url{https://qwenlm.ai}, 2025.
\newblock Large Language Model.

\bibitem[Antonelli et~al.(2022)Antonelli, Reinke, Bakas, Farahani, Kopp-Schneider, Landman, Litjens, Menze, Ronneberger, Summers, van Ginneken, Bilello, Bilic, Christ, Do, Gollub, Heckers, Huisman, Jarnagin, McHugo, Napel, Pernicka, Rhode, Tobon-Gomez, Vorontsov, Meakin, Ourselin, Wiesenfarth, Arbeláez, Bae, Chen, Daza, Feng, He, Isensee, Ji, Jia, Kim, Maier-Hein, Merhof, Pai, Park, Perslev, Rezaiifar, Rippel, Sarasua, Shen, Son, Wachinger, Wang, Wang, Xia, Xu, Xu, Zheng, Simpson, Maier-Hein, and Cardoso]{Antonelli_2022}
Michela Antonelli, Annika Reinke, Spyridon Bakas, Keyvan Farahani, Annette Kopp-Schneider, Bennett~A. Landman, Geert Litjens, Bjoern Menze, Olaf Ronneberger, Ronald~M. Summers, Bram van Ginneken, Michel Bilello, Patrick Bilic, Patrick~F. Christ, Richard K.~G. Do, Marc~J. Gollub, Stephan~H. Heckers, Henkjan Huisman, William~R. Jarnagin, Maureen~K. McHugo, Sandy Napel, Jennifer S.~Golia Pernicka, Kawal Rhode, Catalina Tobon-Gomez, Eugene Vorontsov, James~A. Meakin, Sebastien Ourselin, Manuel Wiesenfarth, Pablo Arbeláez, Byeonguk Bae, Sihong Chen, Laura Daza, Jianjiang Feng, Baochun He, Fabian Isensee, Yuanfeng Ji, Fucang Jia, Ildoo Kim, Klaus Maier-Hein, Dorit Merhof, Akshay Pai, Beomhee Park, Mathias Perslev, Ramin Rezaiifar, Oliver Rippel, Ignacio Sarasua, Wei Shen, Jaemin Son, Christian Wachinger, Liansheng Wang, Yan Wang, Yingda Xia, Daguang Xu, Zhanwei Xu, Yefeng Zheng, Amber~L. Simpson, Lena Maier-Hein, and M.~Jorge Cardoso.
\newblock The medical segmentation decathlon.
\newblock \emph{Nature Communications}, 13\penalty0 (1), 2022.

\bibitem[Avants et~al.(2011)Avants, Tustison, Song, Cook, Klein, and Gee]{avants2011reproducible}
Brian~B Avants, Nicholas~J Tustison, Gang Song, Philip~A Cook, Arno Klein, and James~C Gee.
\newblock A reproducible evaluation of ants similarity metric performance in brain image registration.
\newblock \emph{Neuroimage}, 54\penalty0 (3):\penalty0 2033--2044, 2011.

\bibitem[Bartnik et~al.(2024)Bartnik, Bartczak, Krzyzi{\'n}ski, Korzeniowski, Lamparski, W\k{e}grzyn, Lam, Bartkowiak, Wr{\'o}blewski, Mech, Januszewicz, and Biecek]{doi:10.1148/ryai.240296}
Krzysztof Bartnik, Tomasz Bartczak, Mateusz Krzyzi{\'n}ski, Krzysztof Korzeniowski, Krzysztof Lamparski, Piotr W\k{e}grzyn, Eric Lam, Mateusz Bartkowiak, Tadeusz Wr{\'o}blewski, Katarzyna Mech, Magdalena Januszewicz, and Przemys\l~aw Biecek.
\newblock Waw-tace: A hepatocellular carcinoma multiphase ct dataset with segmentations, radiomics features, and clinical data.
\newblock \emph{Radiology: Artificial Intelligence}, 6\penalty0 (6):\penalty0 e240296, 2024.
\newblock PMID: 39441110.

\bibitem[Bassi et~al.(2024)Bassi, Li, Tang, Isensee, Wang, Chen, Chou, Kirchhoff, Rokuss, Huang, Ye, He, Wald, Ulrich, Baumgartner, Roy, Maier-Hein, Jaeger, Ye, Xie, Zhang, Chen, Xia, Xing, Zhu, Sadegheih, Bozorgpour, Kumari, Azad, Merhof, Shi, Ma, Du, Bai, Huang, Zhao, Wang, Li, Gu, Dong, Yang, Mazurowski, Gupta, Wu, Zhuang, Chen, Roth, Xu, Blaschko, Decherchi, Cavalli, Yuille, and Zhou]{bassi2024touchstone}
Pedro~RAS Bassi, Wenxuan Li, Yucheng Tang, Fabian Isensee, Zifu Wang, Jieneng Chen, Yu-Cheng Chou, Yannick Kirchhoff, Maximilian Rokuss, Ziyan Huang, Jin Ye, Junjun He, Tassilo Wald, Constantin Ulrich, Michael Baumgartner, Saikat Roy, Klaus~H. Maier-Hein, Paul Jaeger, Yiwen Ye, Yutong Xie, Jianpeng Zhang, Ziyang Chen, Yong Xia, Zhaohu Xing, Lei Zhu, Yousef Sadegheih, Afshin Bozorgpour, Pratibha Kumari, Reza Azad, Dorit Merhof, Pengcheng Shi, Ting Ma, Yuxin Du, Fan Bai, Tiejun Huang, Bo Zhao, Haonan Wang, Xiaomeng Li, Hanxue Gu, Haoyu Dong, Jichen Yang, Maciej~A. Mazurowski, Saumya Gupta, Linshan Wu, Jiaxin Zhuang, Hao Chen, Holger Roth, Daguang Xu, Matthew~B. Blaschko, Sergio Decherchi, Andrea Cavalli, Alan~L. Yuille, and Zongwei Zhou.
\newblock Touchstone benchmark: Are we on the right way for evaluating ai algorithms for medical segmentation?
\newblock \emph{Conference on Neural Information Processing Systems}, 2024.

\bibitem[Bassi et~al.(2025{\natexlab{a}})Bassi, Yavuz, Hamamci, Er, Chen, Li, Menze, Decherchi, Cavalli, Wang, Yang, Yuille, and Zhou]{bassi2025radgpt}
Pedro~RAS Bassi, Mehmet~Can Yavuz, Ibrahim~Ethem Hamamci, Sezgin Er, Xiaoxi Chen, Wenxuan Li, Bjoern Menze, Sergio Decherchi, Andrea Cavalli, Kang Wang, Yang Yang, Alan Yuille, and Zongwei Zhou.
\newblock Radgpt: Constructing 3d image-text tumor datasets.
\newblock In \emph{Proceedings of the IEEE/CVF International Conference on Computer Vision}, pages 23720--23730, 2025{\natexlab{a}}.

\bibitem[Bassi et~al.(2025{\natexlab{b}})Bassi, Zhou, Li, P{\l}otka, Chen, Chen, Zhu, Prz{\k{a}}do, Hamamci, Er, Chen, Yavuz, Chou, Lin, Wang, Tang, Cwikla, Decherchi, Cavalli, Yang, Yuille, and Zhou]{bassi2025scaling}
Pedro~RAS Bassi, Xinze Zhou, Wenxuan Li, Szymon P{\l}otka, Jieneng Chen, Qi Chen, Zheren Zhu, Jakub Prz{\k{a}}do, Ibrahim~E Hamamci, Sezgin Er, Xiaoxi Chen, Mehmet~Can Yavuz, Yu-Cheng Chou, Tianyu Lin, Kang Wang, Yucheng Tang, Jaroslaw~B Cwikla, Sergio Decherchi, Andrea Cavalli, Yang Yang, Alan~L Yuille, and Zongwei Zhou.
\newblock Scaling artificial intelligence for multi-tumor early detection with more reports, fewer masks.
\newblock \emph{arXiv preprint arXiv:2510.14803}, 2025{\natexlab{b}}.

\bibitem[Bilic et~al.(2023)Bilic, Christ, Li, Vorontsov, Ben-Cohen, Kaissis, Szeskin, Jacobs, Mamani, Chartrand, Lohöfer, Holch, Sommer, Hofmann, Hostettler, Lev-Cohain, Drozdzal, Amitai, Vivanti, Sosna, Ezhov, Sekuboyina, Navarro, Kofler, Paetzold, Shit, Hu, Lipková, Rempfler, Piraud, Kirschke, Wiestler, Zhang, Hülsemeyer, Beetz, Ettlinger, Antonelli, Bae, Bellver, Bi, Chen, Chlebus, Dam, Dou, Fu, Georgescu, i~Nieto, Gruen, Han, Heng, Hesser, Moltz, Igel, Isensee, Jäger, Jia, Kaluva, Khened, Kim, Kim, Kim, Kohl, Konopczynski, Kori, Krishnamurthi, Li, Li, Li, Li, Lowengrub, Ma, Maier-Hein, Maninis, Meine, Merhof, Pai, Perslev, Petersen, Pont-Tuset, Qi, Qi, Rippel, Roth, Sarasua, Schenk, Shen, Torres, Wachinger, Wang, Weninger, Wu, Xu, Yang, Yu, Yuan, Yue, Zhang, Cardoso, Bakas, Braren, Heinemann, Pal, Tang, Kadoury, Soler, {van Ginneken}, Greenspan, Joskowicz, and Menze]{BILIC2023102680}
Patrick Bilic, Patrick Christ, Hongwei~Bran Li, Eugene Vorontsov, Avi Ben-Cohen, Georgios Kaissis, Adi Szeskin, Colin Jacobs, Gabriel Efrain~Humpire Mamani, Gabriel Chartrand, Fabian Lohöfer, Julian~Walter Holch, Wieland Sommer, Felix Hofmann, Alexandre Hostettler, Naama Lev-Cohain, Michal Drozdzal, Michal~Marianne Amitai, Refael Vivanti, Jacob Sosna, Ivan Ezhov, Anjany Sekuboyina, Fernando Navarro, Florian Kofler, Johannes~C. Paetzold, Suprosanna Shit, Xiaobin Hu, Jana Lipková, Markus Rempfler, Marie Piraud, Jan Kirschke, Benedikt Wiestler, Zhiheng Zhang, Christian Hülsemeyer, Marcel Beetz, Florian Ettlinger, Michela Antonelli, Woong Bae, Míriam Bellver, Lei Bi, Hao Chen, Grzegorz Chlebus, Erik~B. Dam, Qi Dou, Chi-Wing Fu, Bogdan Georgescu, Xavier~Giró i Nieto, Felix Gruen, Xu Han, Pheng-Ann Heng, Jürgen Hesser, Jan~Hendrik Moltz, Christian Igel, Fabian Isensee, Paul Jäger, Fucang Jia, Krishna~Chaitanya Kaluva, Mahendra Khened, Ildoo Kim, Jae-Hun Kim, Sungwoong Kim, Simon Kohl, Tomasz Konopczynski,
  Avinash Kori, Ganapathy Krishnamurthi, Fan Li, Hongchao Li, Junbo Li, Xiaomeng Li, John Lowengrub, Jun Ma, Klaus Maier-Hein, Kevis-Kokitsi Maninis, Hans Meine, Dorit Merhof, Akshay Pai, Mathias Perslev, Jens Petersen, Jordi Pont-Tuset, Jin Qi, Xiaojuan Qi, Oliver Rippel, Karsten Roth, Ignacio Sarasua, Andrea Schenk, Zengming Shen, Jordi Torres, Christian Wachinger, Chunliang Wang, Leon Weninger, Jianrong Wu, Daguang Xu, Xiaoping Yang, Simon Chun-Ho Yu, Yading Yuan, Miao Yue, Liping Zhang, Jorge Cardoso, Spyridon Bakas, Rickmer Braren, Volker Heinemann, Christopher Pal, An Tang, Samuel Kadoury, Luc Soler, Bram {van Ginneken}, Hayit Greenspan, Leo Joskowicz, and Bjoern Menze.
\newblock The liver tumor segmentation benchmark (lits).
\newblock \emph{Medical Image Analysis}, 84:\penalty0 102680, 2023.

\bibitem[Cai et~al.(2024{\natexlab{a}})Cai, Liang, Wang, Wang, Zhang, Yang, Zhou, and Yuille]{cai2024radiative}
Yuanhao Cai, Yixun Liang, Jiahao Wang, Angtian Wang, Yulun Zhang, Xiaokang Yang, Zongwei Zhou, and Alan Yuille.
\newblock Radiative gaussian splatting for efficient x-ray novel view synthesis.
\newblock \emph{arXiv preprint arXiv:2403.04116}, 2024{\natexlab{a}}.

\bibitem[Cai et~al.(2024{\natexlab{b}})Cai, Wang, Yuille, Zhou, and Wang]{cai2024structure}
Yuanhao Cai, Jiahao Wang, Alan Yuille, Zongwei Zhou, and Angtian Wang.
\newblock Structure-aware sparse-view x-ray 3d reconstruction.
\newblock In \emph{IEEE/CVF Conference on Computer Vision and Pattern Recognition (CVPR)}, pages 11174--11183, 2024{\natexlab{b}}.

\bibitem[Cao et~al.(2023)Cao, Xia, Yao, Han, Lambert, Zhang, Tang, Jin, Jiang, Fang, et~al.]{cao2023large}
Kai Cao, Yingda Xia, Jiawen Yao, Xu Han, Lukas Lambert, Tingting Zhang, Wei Tang, Gang Jin, Hui Jiang, Xu Fang, et~al.
\newblock Large-scale pancreatic cancer detection via non-contrast ct and deep learning.
\newblock \emph{Nature medicine}, 29\penalty0 (12):\penalty0 3033--3043, 2023.

\bibitem[Chen et~al.(2024{\natexlab{a}})Chen, Chen, Song, Xiong, Yuille, Wei, and Zhou]{chen2024towards}
Qi Chen, Xiaoxi Chen, Haorui Song, Zhiwei Xiong, Alan Yuille, Chen Wei, and Zongwei Zhou.
\newblock Towards generalizable tumor synthesis.
\newblock In \emph{IEEE/CVF conference on computer vision and pattern recognition (CVPR)}, pages 11147--11158, 2024{\natexlab{a}}.

\bibitem[Chen et~al.(2024{\natexlab{b}})Chen, Lai, Chen, Hu, Yuille, and Zhou]{chen2024analyzing}
Qi Chen, Yuxiang Lai, Xiaoxi Chen, Qixin Hu, Alan Yuille, and Zongwei Zhou.
\newblock Analyzing tumors by synthesis.
\newblock \emph{Generative Machine Learning Models in Medical Image Computing}, pages 85--110, 2024{\natexlab{b}}.

\bibitem[Chen et~al.(2025{\natexlab{a}})Chen, Zhou, Liu, Chen, Li, Jiang, Huang, Zhao, Yu, He, Zheng, Shao, Yuille, and Zhou]{chen2025scaling}
Qi Chen, Xinze Zhou, Chen Liu, Hao Chen, Wenxuan Li, Zekun Jiang, Ziyan Huang, Yuxuan Zhao, Dexin Yu, Junjun He, Yefeng Zheng, Ling Shao, Alan Yuille, and Zongwei Zhou.
\newblock Scaling tumor segmentation: Best lessons from real and synthetic data.
\newblock In \emph{Proceedings of the IEEE/CVF International Conference on Computer Vision}, pages 24001--24013, 2025{\natexlab{a}}.

\bibitem[Chen et~al.(2024{\natexlab{c}})Chen, Konz, Gu, Dong, Chen, Li, Lee, and Mazurowski]{chen2024contourdiff}
Yuwen Chen, Nicholas Konz, Hanxue Gu, Haoyu Dong, Yaqian Chen, Lin Li, Jisoo Lee, and Maciej~A Mazurowski.
\newblock Contourdiff: Unpaired image-to-image translation with structural consistency for medical imaging.
\newblock \emph{arXiv preprint arXiv:2403.10786}, 2024{\natexlab{c}}.

\bibitem[Chen et~al.(2025{\natexlab{b}})Chen, Xiao, Bassi, Zhou, Er, Hamamci, Zhou, and Yuille]{chen2025vision}
Yixiong Chen, Wenjie Xiao, Pedro~RAS Bassi, Xinze Zhou, Sezgin Er, Ibrahim~Ethem Hamamci, Zongwei Zhou, and Alan Yuille.
\newblock Are vision language models ready for clinical diagnosis? a 3d medical benchmark for tumor-centric visual question answering.
\newblock \emph{arXiv preprint arXiv:2505.18915}, 2025{\natexlab{b}}.

\bibitem[Chou et~al.(2024)Chou, Zhou, and Yuille]{chou2024embracing}
Yu-Cheng Chou, Zongwei Zhou, and Alan Yuille.
\newblock Embracing massive medical data.
\newblock In \emph{International Conference on Medical Image Computing and Computer-Assisted Intervention}, pages 24--35. Springer, 2024.

\bibitem[Chu et~al.(2017)Chu, Zhmoginov, and Sandler]{chu2017cyclegan}
Casey Chu, Andrey Zhmoginov, and Mark Sandler.
\newblock Cyclegan, a master of steganography.
\newblock \emph{arXiv preprint arXiv:1712.02950}, 2017.

\bibitem[Dao et~al.(2022)Dao, Nguyen, Pham, and Nguyen]{dao2022phase}
Binh~T Dao, Thang~V Nguyen, Hieu~H Pham, and Ha~Q Nguyen.
\newblock Phase recognition in contrast-enhanced ct scans based on deep learning and random sampling.
\newblock \emph{Medical Physics}, 49\penalty0 (7):\penalty0 4518--4528, 2022.

\bibitem[DenOtter and Schubert(2019)]{denotter2019hounsfield}
Tami~D DenOtter and Johanna Schubert.
\newblock Hounsfield unit.
\newblock 2019.

\bibitem[Du et~al.(2024)Du, Wang, Lu, Zhou, Zhang, Yuille, Li, and Zhou]{du2024boosting}
Shiyi Du, Xiaosong Wang, Yongyi Lu, Yuyin Zhou, Shaoting Zhang, Alan Yuille, Kang Li, and Zongwei Zhou.
\newblock Boosting dermatoscopic lesion segmentation via diffusion models with visual and textual prompts.
\newblock In \emph{2024 IEEE International Symposium on Biomedical Imaging (ISBI)}, pages 1--5. IEEE, 2024.

\bibitem[Esser et~al.(2021)Esser, Rombach, and Ommer]{esser2021taming}
Patrick Esser, Robin Rombach, and Bjorn Ommer.
\newblock Taming transformers for high-resolution image synthesis.
\newblock In \emph{Proceedings of the IEEE/CVF conference on computer vision and pattern recognition}, pages 12873--12883, 2021.

\bibitem[Gao et~al.(2022)Gao, Zhou, Liu, Yan, Zhang, and Metaxas]{gao2022data}
Yunhe Gao, Mu Zhou, Di Liu, Zhennan Yan, Shaoting Zhang, and Dimitris~N Metaxas.
\newblock A data-scalable transformer for medical image segmentation: architecture, model efficiency, and benchmark.
\newblock \emph{arXiv preprint arXiv:2203.00131}, 2022.

\bibitem[Gardezi et~al.(2024)Gardezi, Aronson, Wawrzyn, Yu, Abel, Shapiro, Lubner, Warner, Toia, Mao, et~al.]{gardezi2024resnct}
Syed Jamal~Safdar Gardezi, Lucas Aronson, Peter Wawrzyn, Hongkun Yu, E~Jason Abel, Daniel~D Shapiro, Meghan~G Lubner, Joshua Warner, Giuseppe Toia, Lu Mao, et~al.
\newblock Resnct: A deep learning model for the synthesis of nephrographic phase images in ct urography.
\newblock \emph{arXiv preprint arXiv:2405.04629}, 2024.

\bibitem[{Google DeepMind}(2025)]{google_nanobanana}
{Google DeepMind}.
\newblock Gemini nano banana.
\newblock \url{https://deepmind.google}, 2025.
\newblock Lightweight Multimodal Model.

\bibitem[Guo et~al.(2025)Guo, Zhao, Yang, He, Nath, Xu, Bassi, Zhou, Simon, Harmon, Syed, Roth, and Xu]{guo2025text2ct}
Pengfei Guo, Can Zhao, Dong Yang, Yufan He, Vishwesh Nath, Ziyue Xu, Pedro~RAS Bassi, Zongwei Zhou, Benjamin~D Simon, Stephanie~Anne Harmon, Ali~B Syed, Holger Roth, and Daguang Xu.
\newblock Text2ct: Towards 3d ct volume generation from free-text descriptions using diffusion model.
\newblock \emph{arXiv preprint arXiv:2505.04522}, 2025.

\bibitem[Hamamci et~al.(2025)Hamamci, Er, Wang, Almas, Simsek, Esirgun, Dogan, Durugol, Hou, Shit, Dai, Xu, Reynaud, Dasdelen, Wittmann, Amiranashvili, Simsar, Simsar, Erdemir, Alanbay, Sekuboyina, Lafci, Kaplan, Lu, Polacin, Kainz, Bluethgen, Batmanghelich, Ozdemir, and Menze]{hamamci2025developinggeneralistfoundationmodels}
Ibrahim~Ethem Hamamci, Sezgin Er, Chenyu Wang, Furkan Almas, Ayse~Gulnihan Simsek, Sevval~Nil Esirgun, Irem Dogan, Omer~Faruk Durugol, Benjamin Hou, Suprosanna Shit, Weicheng Dai, Murong Xu, Hadrien Reynaud, Muhammed~Furkan Dasdelen, Bastian Wittmann, Tamaz Amiranashvili, Enis Simsar, Mehmet Simsar, Emine~Bensu Erdemir, Abdullah Alanbay, Anjany Sekuboyina, Berkan Lafci, Ahmet Kaplan, Zhiyong Lu, Malgorzata Polacin, Bernhard Kainz, Christian Bluethgen, Kayhan Batmanghelich, Mehmet~Kemal Ozdemir, and Bjoern Menze.
\newblock Developing generalist foundation models from a multimodal dataset for 3d computed tomography, 2025.

\bibitem[He et~al.(2025)He, Guo, Tang, Myronenko, Nath, Xu, Yang, Zhao, Simon, Belue, et~al.]{he2025vista3d}
Yufan He, Pengfei Guo, Yucheng Tang, Andriy Myronenko, Vishwesh Nath, Ziyue Xu, Dong Yang, Can Zhao, Benjamin Simon, Mason Belue, et~al.
\newblock Vista3d: A unified segmentation foundation model for 3d medical imaging.
\newblock In \emph{Proceedings of the Computer Vision and Pattern Recognition Conference}, pages 20863--20873, 2025.

\bibitem[Heusel et~al.(2017)Heusel, Ramsauer, Unterthiner, Nessler, and Hochreiter]{heusel2017gans}
Martin Heusel, Hubert Ramsauer, Thomas Unterthiner, Bernhard Nessler, and Sepp Hochreiter.
\newblock Gans trained by a two time-scale update rule converge to a local nash equilibrium.
\newblock \emph{Advances in neural information processing systems}, 30, 2017.

\bibitem[Ho et~al.(2020)Ho, Jain, and Abbeel]{ho2020denoising}
Jonathan Ho, Ajay Jain, and Pieter Abbeel.
\newblock Denoising diffusion probabilistic models.
\newblock \emph{Advances in neural information processing systems}, 33:\penalty0 6840--6851, 2020.

\bibitem[Hu et~al.(2025)Hu, Xia, Zheng, Cao, Zheng, Chen, Sun, Chen, Zheng, Pan, et~al.]{hu2025ai}
Can Hu, Yingda Xia, Zhilin Zheng, Mengxuan Cao, Guoliang Zheng, Shangqi Chen, Jiancheng Sun, Wujie Chen, Qi Zheng, Siwei Pan, et~al.
\newblock Ai-based large-scale screening of gastric cancer from noncontrast ct imaging.
\newblock \emph{Nature Medicine}, pages 1--9, 2025.

\bibitem[Hu et~al.(2022)Hu, Xiao, Chen, Sun, Chen, Yuille, and Zhou]{hu2022synthetic}
Qixin Hu, Junfei Xiao, Yixiong Chen, Shuwen Sun, Jie-Neng Chen, Alan Yuille, and Zongwei Zhou.
\newblock Synthetic tumors make ai segment tumors better.
\newblock \emph{NeurIPS Workshop on Medical Imaging meets NeurIPS}, 2022.

\bibitem[Hu et~al.(2023{\natexlab{a}})Hu, Chen, Xiao, Sun, Chen, Yuille, and Zhou]{hu2023label}
Qixin Hu, Yixiong Chen, Junfei Xiao, Shuwen Sun, Jieneng Chen, Alan~L Yuille, and Zongwei Zhou.
\newblock Label-free liver tumor segmentation.
\newblock In \emph{IEEE/CVF Conference on Computer Vision and Pattern Recognition (CVPR)}, pages 7422--7432, 2023{\natexlab{a}}.

\bibitem[Hu et~al.(2023{\natexlab{b}})Hu, Yuille, and Zhou]{hu2023synthetic}
Qixin Hu, Alan Yuille, and Zongwei Zhou.
\newblock Synthetic data as validation.
\newblock \emph{arXiv preprint arXiv:2310.16052}, 2023{\natexlab{b}}.

\bibitem[Huang et~al.(2017)Huang, Liu, Van Der~Maaten, and Weinberger]{huang2017densely}
Gao Huang, Zhuang Liu, Laurens Van Der~Maaten, and Kilian~Q Weinberger.
\newblock Densely connected convolutional networks.
\newblock In \emph{Proceedings of the IEEE conference on computer vision and pattern recognition}, pages 4700--4708, 2017.

\bibitem[Huang et~al.(2018)Huang, Liu, Belongie, and Kautz]{huang2018multimodal}
Xun Huang, Ming-Yu Liu, Serge Belongie, and Jan Kautz.
\newblock Multimodal unsupervised image-to-image translation.
\newblock In \emph{Proceedings of the European conference on computer vision (ECCV)}, pages 172--189, 2018.

\bibitem[Isensee et~al.(2021)Isensee, Jaeger, Kohl, Petersen, and Maier-Hein]{isensee2021nnu}
Fabian Isensee, Paul~F Jaeger, Simon~AA Kohl, Jens Petersen, and Klaus~H Maier-Hein.
\newblock nnu-net: a self-configuring method for deep learning-based biomedical image segmentation.
\newblock \emph{Nature methods}, 18\penalty0 (2):\penalty0 203--211, 2021.

\bibitem[Isola et~al.(2017)Isola, Zhu, Zhou, and Efros]{isola2017image}
Phillip Isola, Jun-Yan Zhu, Tinghui Zhou, and Alexei~A Efros.
\newblock Image-to-image translation with conditional adversarial networks.
\newblock In \emph{Proceedings of the IEEE conference on computer vision and pattern recognition}, pages 1125--1134, 2017.

\bibitem[Khader et~al.(2023)Khader, M{\"u}ller-Franzes, Tayebi~Arasteh, Han, Haarburger, Schulze-Hagen, Schad, Engelhardt, Bae{\ss}ler, Foersch, et~al.]{khader2023denoising}
Firas Khader, Gustav M{\"u}ller-Franzes, Soroosh Tayebi~Arasteh, Tianyu Han, Christoph Haarburger, Maximilian Schulze-Hagen, Philipp Schad, Sandy Engelhardt, Bettina Bae{\ss}ler, Sebastian Foersch, et~al.
\newblock Denoising diffusion probabilistic models for 3d medical image generation.
\newblock \emph{Scientific Reports}, 13\penalty0 (1):\penalty0 7303, 2023.

\bibitem[Kirk et~al.(2016)Kirk, Lee, Lucchesi, Aredes, Gruszauskas, Catto, Garcia, Jarosz, Duddalwar, Varghese, Rieger-Christ, and Lemmerman]{Kirk2016_TCGA-BLCA}
S. Kirk, Y. Lee, F.~R. Lucchesi, N.~D. Aredes, N. Gruszauskas, J. Catto, K. Garcia, R. Jarosz, V. Duddalwar, B. Varghese, K. Rieger-Christ, and J. Lemmerman.
\newblock The cancer genome atlas urothelial bladder carcinoma collection (tcga-blca) (version 8) [data set].
\newblock The Cancer Imaging Archive, 2016.
\newblock Retrieved from https://doi.org/10.7937/K9/TCIA.2016.8LNG8XDR.

\bibitem[Korhonen and You(2012)]{korhonen2012peak}
Jari Korhonen and Junyong You.
\newblock Peak signal-to-noise ratio revisited: Is simple beautiful?
\newblock In \emph{2012 Fourth international workshop on quality of multimedia experience}, pages 37--38. IEEE, 2012.

\bibitem[Kwon and Ye(2022)]{kwon2022diffusion}
Gihyun Kwon and Jong~Chul Ye.
\newblock Diffusion-based image translation using disentangled style and content representation.
\newblock \emph{arXiv preprint arXiv:2209.15264}, 2022.

\bibitem[Lai et~al.(2024)Lai, Chen, Wang, Yuille, and Zhou]{lai2024pixel}
Yuxiang Lai, Xiaoxi Chen, Angtian Wang, Alan Yuille, and Zongwei Zhou.
\newblock From pixel to cancer: Cellular automata in computed tomography.
\newblock In \emph{International Conference on Medical Image Computing and Computer-Assisted Intervention (MICCAI)}, pages 36--46. Springer, 2024.

\bibitem[Li et~al.(2023)Li, Chou, Sun, Qiao, Yuille, and Zhou]{li2023early}
Bowen Li, Yu-Cheng Chou, Shuwen Sun, Hualin Qiao, Alan Yuille, and Zongwei Zhou.
\newblock Early detection and localization of pancreatic cancer by label-free tumor synthesis.
\newblock \emph{MICCAI Workshop on Big Task Small Data, 1001-AI}, 2023.

\bibitem[Li et~al.(2024{\natexlab{a}})Li, Zhou, Yang, Pepe, Gsaxner, Luijten, Qu, Zhang, Chen, Li, Jin, and Egger]{li2024medshapenet}
Jianning Li, Zongwei Zhou, Jiancheng Yang, Antonio Pepe, Christina Gsaxner, Gijs Luijten, Chongyu Qu, Tiezheng Zhang, Xiaoxi Chen, Wenxuan Li, Yuan Jin, and Jan Egger.
\newblock Medshapenet--a large-scale dataset of 3d medical shapes for computer vision.
\newblock \emph{Biomedical Engineering/Biomedizinische Technik}, \penalty0 (0), 2024{\natexlab{a}}.

\bibitem[Li et~al.(2024{\natexlab{b}})Li, Qu, Chen, Bassi, Shi, Lai, Yu, Xue, Chen, Lin, Tang, Cao, Han, Zhang, Liu, Zhang, Ma, Wang, Zhang, Yuille, and Zhou]{li2024abdomenatlas}
Wenxuan Li, Chongyu Qu, Xiaoxi Chen, Pedro~RAS Bassi, Yijia Shi, Yuxiang Lai, Qian Yu, Huimin Xue, Yixiong Chen, Xiaorui Lin, Yutong Tang, Yining Cao, Haoqi Han, Zheyuan Zhang, Jiawei Liu, Tiezheng Zhang, Yujiu Ma, Jincheng Wang, Guang Zhang, Alan Yuille, and Zongwei Zhou.
\newblock Abdomenatlas: A large-scale, detailed-annotated, \& multi-center dataset for efficient transfer learning and open algorithmic benchmarking.
\newblock \emph{Medical Image Analysis}, page 103285, 2024{\natexlab{b}}.

\bibitem[Li et~al.(2024{\natexlab{c}})Li, Yuille, and Zhou]{li2024well}
Wenxuan Li, Alan Yuille, and Zongwei Zhou.
\newblock How well do supervised models transfer to 3d image segmentation?
\newblock In \emph{International Conference on Learning Representations}, 2024{\natexlab{c}}.

\bibitem[Li et~al.(2025{\natexlab{a}})Li, Bassi, Lin, Chou, Zhou, Tang, Isensee, Wang, Chen, Xu, Ye, Zhu, Decherchi, Cavalli, Yuille, and Zhou]{li2025scalemai}
Wenxuan Li, Pedro~RAS Bassi, Tianyu Lin, Yu-Cheng Chou, Xinze Zhou, Yucheng Tang, Fabian Isensee, Kang Wang, Qi Chen, Xiaowei Xu, Jin Ye, Zheren Zhu, Sergio Decherchi, Andrea Cavalli, Alan~L Yuille, and Zongwei Zhou.
\newblock Scalemai: Accelerating the development of trusted datasets and ai models.
\newblock \emph{arXiv preprint arXiv:2501.03410}, 2025{\natexlab{a}}.

\bibitem[Li et~al.(2025{\natexlab{b}})Li, Zhou, Chen, Lin, Bassi, Plotka, Cwikla, Chen, Ye, Zhu, Chou, Wang, Tang, Yuille, and Zhou]{li2025pants}
Wenxuan Li, Xinze Zhou, Qi Chen, Tianyu Lin, Pedro~RAS Bassi, Szymon Plotka, Jaroslaw~B Cwikla, Xiaoxi Chen, Chen Ye, Zheren Zhu, Yu-Cheng Chou, Kang Wang, Yucheng Tang, Alan~L Yuille, and Zongwei Zhou.
\newblock Pants: The pancreatic tumor segmentation dataset.
\newblock \emph{arXiv preprint arXiv:2507.01291}, 2025{\natexlab{b}}.

\bibitem[Li et~al.(2024{\natexlab{d}})Li, Shuai, Liu, Chen, Wu, Guo, Yang, Zhao, Bassi, Xu, and Zhou]{li2024text}
Xinran Li, Yi Shuai, Chen Liu, Qi Chen, Qilong Wu, Pengfei Guo, Dong Yang, Can Zhao, Pedro~RAS Bassi, Daguang Xu, and Zongwei Zhou.
\newblock Text-driven tumor synthesis.
\newblock \emph{arXiv preprint arXiv:2412.18589}, 2024{\natexlab{d}}.

\bibitem[Lin et~al.(2025)Lin, Li, Zhuang, Chen, Cai, Ding, Yuille, and Zhou]{lin2025pixel}
Tianyu Lin, Xinran Li, Chuntung Zhuang, Qi Chen, Yuanhao Cai, Kai Ding, Alan~L Yuille, and Zongwei Zhou.
\newblock Are pixel-wise metrics reliable for sparse-view computed tomography reconstruction?
\newblock \emph{arXiv preprint arXiv:2506.02093}, 2025.

\bibitem[Liu et~al.(2023)Liu, Zhang, Chen, Xiao, Lu, Landman, Yuan, Yuille, Tang, and Zhou]{liu2023clip}
Jie Liu, Yixiao Zhang, Jie-Neng Chen, Junfei Xiao, Yongyi Lu, Bennett~A Landman, Yixuan Yuan, Alan Yuille, Yucheng Tang, and Zongwei Zhou.
\newblock Clip-driven universal model for organ segmentation and tumor detection.
\newblock In \emph{Proceedings of the IEEE/CVF International Conference on Computer Vision}, pages 21152--21164, 2023.

\bibitem[Liu et~al.(2024)Liu, Zhang, Wang, Yavuz, Chen, Yuan, Li, Yang, Yuille, Tang, and Zhou]{liu2024universal}
Jie Liu, Yixiao Zhang, Kang Wang, Mehmet~Can Yavuz, Xiaoxi Chen, Yixuan Yuan, Haoliang Li, Yang Yang, Alan Yuille, Yucheng Tang, and Zongwei Zhou.
\newblock Universal and extensible language-vision models for organ segmentation and tumor detection from abdominal computed tomography.
\newblock \emph{Medical Image Analysis}, page 103226, 2024.

\bibitem[Luo et~al.(2024)Luo, Yang, Liu, Shi, Huang, Zheng, and Cheng]{luo2024target}
Yimin Luo, Qinyu Yang, Ziyi Liu, Zenglin Shi, Weimin Huang, Guoyan Zheng, and Jun Cheng.
\newblock Target-guided diffusion models for unpaired cross-modality medical image translation.
\newblock \emph{IEEE Journal of Biomedical and Health Informatics}, 28\penalty0 (7):\penalty0 4062--4071, 2024.

\bibitem[Mao et~al.(2025)Mao, Wang, Tang, Xu, Wang, Yang, Zhou, and Zhou]{mao2025medsegfactory}
Jiawei Mao, Yuhan Wang, Yucheng Tang, Daguang Xu, Kang Wang, Yang Yang, Zongwei Zhou, and Yuyin Zhou.
\newblock Medsegfactory: Text-guided generation of medical image-mask pairs.
\newblock \emph{arXiv preprint arXiv:2504.06897}, 2025.

\bibitem[{OpenAI}(2025)]{chatgpt51}
{OpenAI}.
\newblock Chatgpt-5.1.
\newblock \url{https://openai.com}, 2025.
\newblock Large Language Model.

\bibitem[{\"O}zbey et~al.(2023){\"O}zbey, Dalmaz, Dar, Bedel, {\"O}zturk, G{\"u}ng{\"o}r, and Cukur]{ozbey2023unsupervised}
Muzaffer {\"O}zbey, Onat Dalmaz, Salman~UH Dar, Hasan~A Bedel, {\c{S}}aban {\"O}zturk, Alper G{\"u}ng{\"o}r, and Tolga Cukur.
\newblock Unsupervised medical image translation with adversarial diffusion models.
\newblock \emph{IEEE Transactions on Medical Imaging}, 42\penalty0 (12):\penalty0 3524--3539, 2023.

\bibitem[Park et~al.(2020)Park, Efros, Zhang, and Zhu]{park2020contrastive}
Taesung Park, Alexei~A Efros, Richard Zhang, and Jun-Yan Zhu.
\newblock Contrastive learning for unpaired image-to-image translation.
\newblock In \emph{European conference on computer vision}, pages 319--345. Springer, 2020.

\bibitem[Phan et~al.(2024)Phan, Xie, Zhang, Qi, Liao, Perperidis, Phung, Verjans, and To]{phan2024structural}
Vu~Minh~Hieu Phan, Yutong Xie, Bowen Zhang, Yuankai Qi, Zhibin Liao, Antonios Perperidis, Son~Lam Phung, Johan~W Verjans, and Minh-Son To.
\newblock Structural attention: Rethinking transformer for unpaired medical image synthesis.
\newblock In \emph{International Conference on Medical Image Computing and Computer-Assisted Intervention}, pages 690--700. Springer, 2024.

\bibitem[Qu et~al.(2023)Qu, Zhang, Qiao, Liu, Tang, Yuille, and Zhou]{qu2023annotating}
Chongyu Qu, Tiezheng Zhang, Hualin Qiao, Jie Liu, Yucheng Tang, Alan Yuille, and Zongwei Zhou.
\newblock Abdomenatlas-8k: Annotating 8,000 abdominal ct volumes for multi-organ segmentation in three weeks.
\newblock In \emph{Conference on Neural Information Processing Systems}, 2023.

\bibitem[Ristea et~al.(2023)Ristea, Miron, Savencu, Georgescu, Verga, Khan, and Ionescu]{ristea2023cytran}
Nicolae-C{\u{a}}t{\u{a}}lin Ristea, Andreea-Iuliana Miron, Olivian Savencu, Mariana-Iuliana Georgescu, Nicolae Verga, Fahad~Shahbaz Khan, and Radu~Tudor Ionescu.
\newblock Cytran: A cycle-consistent transformer with multi-level consistency for non-contrast to contrast ct translation.
\newblock \emph{Neurocomputing}, 538:\penalty0 126211, 2023.

\bibitem[Rombach et~al.(2022)Rombach, Blattmann, Lorenz, Esser, and Ommer]{rombach2022high}
Robin Rombach, Andreas Blattmann, Dominik Lorenz, Patrick Esser, and Bj{\"o}rn Ommer.
\newblock High-resolution image synthesis with latent diffusion models.
\newblock In \emph{Proceedings of the IEEE/CVF conference on computer vision and pattern recognition}, pages 10684--10695, 2022.

\bibitem[Sasaki et~al.(2021)Sasaki, Willcocks, and Breckon]{sasaki2021unit}
Hiroshi Sasaki, Chris~G Willcocks, and Toby~P Breckon.
\newblock Unit-ddpm: Unpaired image translation with denoising diffusion probabilistic models.
\newblock \emph{arXiv preprint arXiv:2104.05358}, 2021.

\bibitem[Siddiquee et~al.(2019)Siddiquee, Zhou, Tajbakhsh, Feng, Gotway, Bengio, and Liang]{siddiquee2019learning}
Md~Mahfuzur~Rahman Siddiquee, Zongwei Zhou, Nima Tajbakhsh, Ruibin Feng, Michael~B Gotway, Yoshua Bengio, and Jianming Liang.
\newblock Learning fixed points in generative adversarial networks: From image-to-image translation to disease detection and localization.
\newblock In \emph{IEEE International Conference on Computer Vision (ICCV)}, pages 191--200, 2019.

\bibitem[Wang et~al.(2024)Wang, Wang, and Cui]{wang2024structure}
Haoshen Wang, Xiaodong Wang, and Zhiming Cui.
\newblock Structure-preserving diffusion model for unpaired medical image translation.
\newblock In \emph{International Workshop on Machine Learning in Medical Imaging}, pages 218--227. Springer, 2024.

\bibitem[Wang et~al.(2004)Wang, Bovik, Sheikh, and Simoncelli]{wang2004image}
Zhou Wang, Alan~C Bovik, Hamid~R Sheikh, and Eero~P Simoncelli.
\newblock Image quality assessment: from error visibility to structural similarity.
\newblock \emph{IEEE transactions on image processing}, 13\penalty0 (4):\penalty0 600--612, 2004.

\bibitem[Xia et~al.(2022)Xia, Yu, Chu, Kawamoto, Park, Liu, Chen, Zhu, Li, Zhou, Yuille, Fishman, and Hruban]{xia2022felix}
Yingda Xia, Qihang Yu, Linda Chu, Satomi Kawamoto, Seyoun Park, Fengze Liu, Jieneng Chen, Zhuotun Zhu, Bowen Li, Zongwei Zhou, Alan~L Yuille, Elliot~K Fishman, and Ralph~H Hruban.
\newblock The felix project: Deep networks to detect pancreatic neoplasms.
\newblock \emph{medRxiv}, 2022.

\bibitem[Xiao et~al.(2022)Xiao, Li, Chen, Zhao, Yang, Xie, Liu, Quan, and Duan]{xiao2022contrast}
Ning Xiao, Zhenyu Li, Shaobo Chen, Liangtian Zhao, Yuer Yang, Hao Xie, Yang Liu, Yujuan Quan, and Junwei Duan.
\newblock Contrast-enhanced ct image synthesis of thyroid based on transfomer and texture branching.
\newblock In \emph{2022 5th International Conference on Artificial Intelligence and Big Data (ICAIBD)}, pages 94--100. IEEE, 2022.

\bibitem[Yang et~al.(2025)Yang, Wang, Liu, Sun, Wang, Chellappa, Zhou, Yuille, Zhu, Zhang, and Chen]{yang2025medical}
Yijun Yang, Zhao-Yang Wang, Qiuping Liu, Shuwen Sun, Kang Wang, Rama Chellappa, Zongwei Zhou, Alan Yuille, Lei Zhu, Yu-Dong Zhang, and Jieneng Chen.
\newblock Medical world model: Generative simulation of tumor evolution for treatment planning.
\newblock \emph{arXiv preprint arXiv:2506.02327}, 2025.

\bibitem[Yin et~al.(2025)Yin, Peng, Li, and Wang]{yin2025enhanced}
Juanjuan Yin, Jinye Peng, Xiaohui Li, and Jun Wang.
\newblock Enhanced aortic ct synthesis based on multiscale information fusion.
\newblock \emph{IEEE MultiMedia}, 2025.

\bibitem[Zhang et~al.(2023)Zhang, Rao, and Agrawala]{zhang2023adding}
Lvmin Zhang, Anyi Rao, and Maneesh Agrawala.
\newblock Adding conditional control to text-to-image diffusion models.
\newblock In \emph{Proceedings of the IEEE/CVF international conference on computer vision}, pages 3836--3847, 2023.

\bibitem[Zhou et~al.(2022)Zhou, Gotway, and Liang]{zhou2022interpreting}
Zongwei Zhou, Michael~B Gotway, and Jianming Liang.
\newblock Interpreting medical images.
\newblock In \emph{Intelligent Systems in Medicine and Health}, pages 343--371. Springer, 2022.

\bibitem[Zhu et~al.(2017)Zhu, Park, Isola, and Efros]{zhu2017unpaired}
Jun-Yan Zhu, Taesung Park, Phillip Isola, and Alexei~A Efros.
\newblock Unpaired image-to-image translation using cycle-consistent adversarial networks.
\newblock In \emph{Proceedings of the IEEE international conference on computer vision}, pages 2223--2232, 2017.

\bibitem[Zhu et~al.()Zhu, Wu, Zhang, and Li]{zhucausality}
Qikui Zhu, Hao Wu, Yanyan Zhang, and Shuo Li.
\newblock Causality-driven spatio-temporal generator for multi-phase contrast-enhanced ct synthesis.

\bibitem[Zhu et~al.(2025{\natexlab{a}})Zhu, Wentland, and Li]{zhu2025contrast}
Qikui Zhu, Andrew~L Wentland, and Shuo Li.
\newblock Contrast-aware network with aggregated-interacted transformer and multi-granularity aligned contrastive learning for synthesizing contrast-enhanced abdomen ct imaging.
\newblock \emph{IEEE Transactions on Computational Imaging}, 2025{\natexlab{a}}.

\bibitem[Zhu et~al.(2025{\natexlab{b}})Zhu, Zhu, Du, and Wang]{zhu2025cross}
Qikui Zhu, Shaoming Zhu, Bo Du, and Yanqing Wang.
\newblock Cross-domain distribution adversarial diffusion model for synthesizing contrast-enhanced abdomen ct imaging.
\newblock \emph{Pattern Recognition}, page 111695, 2025{\natexlab{b}}.

\end{thebibliography}
}

\clearpage
\appendix
\setcounter{page}{1}
\onecolumn
\renewcommand \thepart{}
\renewcommand \partname{}
\part{Appendix} 
\setcounter{secnumdepth}{4}
\setcounter{tocdepth}{4}
\parttoc 

\clearpage

\section{Dataset}\label{sec:supp_dataset}
\subsection{Anatomical Structures in \dataset\ Dataset}
We will present a precisely annotated, phase-wise paired, and organ-wise registered high-quality dataset, \dataset, containing \numofct\ CT volumes from \numofpatient\ patients, collecting from over \numofhospital\ hospitals. For each CT scan, \dataset\ provides 88 anatomical structures.
All volumes were registered at the voxel level using the ANTs~\cite{avants2011reproducible} toolkit to ensure accurate spatial alignment across phases.

\begin{table}[htbp]
\centering
\scriptsize
\caption{Class mapping in the \dataset\ dataset.}
\setlength{\tabcolsep}{4pt}

\begin{tabular*}{\textwidth}{@{\extracolsep{\fill}} lcc lcc lcc lcc}
\toprule
\textbf{name} & type & label &
\textbf{name} & type & label &
\textbf{name} & type & label &
\textbf{name} & type & label \\ 
\midrule

adrenal gland left           & O & 1  &
kidney left                  & O & 23 &
rib left 3                   & B & 45 &
vertebrae C3                 & B & 67 \\

adrenal gland right          & O & 2  &
kidney lesion                & L & 24 &
rib left 4                   & B & 46 &
vertebrae C4                 & B & 68 \\

aorta                        & O & 3  &
kidney right                 & O & 25 &
rib left 5                   & B & 47 &
vertebrae C5                 & B & 69 \\

bile duct (common) & O & 4  &
liver                        & O & 26 &
rib left 6                   & B & 48 &
vertebrae C6                 & B & 70 \\

bladder                      & O & 5  &
liver segment 1              & P & 27 &
rib left 7                   & B & 49 &
vertebrae C7                 & B & 71 \\

celiac trunk                 & O & 6  &
liver segment 2              & P & 28 &
rib left 8                   & B & 50 &
vertebrae L1                 & B & 72 \\

colon                        & O & 7  &
liver segment 3              & P & 29 &
rib left 9                   & B & 51 &
vertebrae L2                 & B & 73 \\

duodenum                     & O & 8  &
liver segment 4              & P & 30 &
rib left 10                  & B & 52 &
vertebrae L3                 & B & 74 \\

esophagus                    & O & 9  &
liver segment 5              & P & 31 &
rib left 11                  & B & 53 &
vertebrae L4                 & B & 75 \\

gall bladder                 & O & 10 &
liver segment 6              & P & 32 &
rib left 12                  & B & 54 &
vertebrae L5                 & B & 76 \\

hepatic vessel               & O & 11 &
liver segment 7              & P & 33 &
rib right 1                  & B & 55 &
vertebrae T1                 & B & 77 \\

intestine                    & O & 12 &
liver segment 8              & P & 34 &
rib right 2                  & B & 56 &
vertebrae T2                 & B & 78 \\

kidney left                  & O & 13 &
liver lesion                 & L & 35 &
rib right 3                  & B & 57 &
vertebrae T3                 & B & 79 \\

kidney right                 & O & 14 &
lung left                    & O & 36 &
rib right 4                  & B & 58 &
vertebrae T4                 & B & 80 \\

kidney lesion                & L & 15 &
lung right                   & O & 37 &
rib right 5                  & B & 59 &
vertebrae T5                 & B & 81 \\

liver                        & O & 16 &
pancreas                     & O & 38 &
rib right 6                  & B & 60 &
vertebrae T6                 & B & 82 \\

liver segment 1              & P & 17 &
pancreas body                & P & 39 &
rib right 7                  & B & 61 &
vertebrae T7                 & B & 83 \\

liver segment 2              & P & 18 &
pancreas head                & P & 40 &
rib right 8                  & B & 62 &
vertebrae T8                 & B & 84 \\

liver segment 3              & P & 19 &
pancreas tail                & P & 41 &
rib right 9                  & B & 63 &
vertebrae T9                 & B & 85 \\

liver segment 4              & P & 20 &
pancreatic lesion            & L & 42 &
rib right 10                 & B & 64 &
vertebrae T10                & B & 86 \\

liver segment 5              & P & 21 &
portal vein and splenic vein & O & 43 &
rib right 11                 & B & 65 &
vertebrae T11                & B & 87 \\

liver segment 6              & P & 22 &
postcava                     & O & 44 &
rib right 12                 & B & 66 &
vertebrae T12                & B & 88 \\

\bottomrule
\end{tabular*}

\begin{tablenotes}
\item type:
\quad \textbf{B}: bone structure 
\quad \textbf{L}: lesion 
\quad \textbf{O}: whole organ 
\quad \textbf{P}: organ sub-region 
\end{tablenotes}
\label{tab: CTVerse}
\end{table}

\subsection{Anatomical Structures in \method\ Test Datasets}
We use these 22 organs to evaluate how well \method\ preserves anatomical structure and tissue characteristics during CT phase conversion.
For each organ, we measure two quantities:
\begin{itemize}
    \item HU (Hounsfield Unit) correlation – whether the tissue density changes follow the correct pattern across phases.
    \item Size correlation – whether the enhanced organ keeps a similar volume as in the real CT.
    
\end{itemize}

\begin{table}[htbp]
\centering
\scriptsize
\caption{Anatomical structure-level HU analysis summary.}
\begin{tabular}{lcc c lcc}
\toprule
name & label & volume & & name & label & volume \\
\midrule

aorta                            & 1  & medium  & &
intestine                        & 12 & large   \\

bladder                          & 2  & medium  & &
kidney left                      & 13 & large   \\

celiac trunk                     & 3  & small   & &
kidney right                     & 14 & large   \\

colon                            & 4  & large   & &
liver                            & 15 & large   \\

duodenum                         & 5  & medium  & &
pancreas body                    & 16 & medium  \\

esophagus                        & 6  & small   & &
pancreas head                    & 17 & medium  \\

gall bladder                     & 7  & small   & &
pancreas tail                    & 18 & small   \\

hepatic vessel                   & 8  & small   & &
portal vein and splenic vein     & 19 & small   \\

prostate                         & 9  & small   & &
postcava                         & 20 & medium  \\

rectum                           & 10 & small   & &
spleen                           & 21 & medium  \\

stomach                          & 11 & large   & &
superior mesenteric artery       & 22 & small   \\
\bottomrule
\end{tabular}
\label{tab:22organ}
\end{table}

\clearpage

\section{CT Contrast Phases}

\subsection{Contrast Phases Change Anatomical Appearance}
A multi-phase CT scan takes several images of the same patient at different times after a contrast agent (iodine) is injected (see \figureautorefname~\ref{fig:supp_multi-phase}).
Think of it like taking four photos of the same scene under different lighting conditions. The anatomy does not change, but the brightness of specific tissues changes over time, because the contrast agent flows through arteries → organs → veins → then slowly washes out.
Each phase highlights different organs and blood vessels, which is important for detecting pancreatic tumors, because they often look very similar to normal tissue in one phase but become visible in another.

\begin{itemize}
    \item \textbf{Non-contrast: This is the scan before contrast injection.} Organs appear in their natural density.
    Pancreatic tumors are often hard to see, because both healthy pancreas and tumors look similar.
    \item  \textbf{Arterial: Taken shortly after contrast agent injection.} Arteries and hyper-vascular structures become bright.
    \item \textbf{Venous: Contrast moves into veins and most abdominal organs.}
    Pancreatic tumors typically appear hypo-enhanced (darker) compared to the enhanced pancreas, making them easier to localize.
    \item \textbf{Delay: Contrast gradually washes out.}
    Tumors may retain different contrast patterns compared to surrounding tissue, providing another chance for detection.
\end{itemize}

\subsection{Contrast Phases Improve Tumor Diagnosis}
Contrast phases play a crucial role in detecting pancreatic tumors because they highlight how different tissues respond to injected contrast over time. In non-contrast scans, tumors often look similar to the surrounding pancreas, making them difficult to spot. Once contrast is injected, however, each phase reveals a different pattern of brightness across organs and vessels.

\begin{figure}[h]
    \centering
    \includegraphics[width=\linewidth]{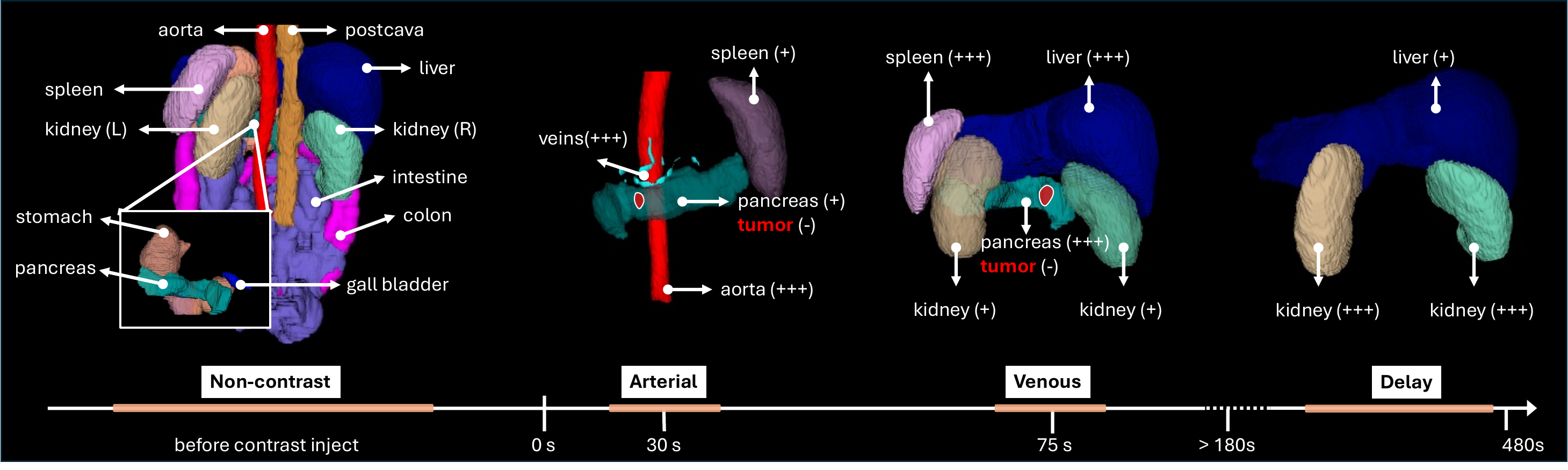}
    \caption{\textbf{Illustration of CT contrast phases.} From left to right: non-contrast phase (with anatomical structure labels), arterial phase, venous phase, and delay phase. In the enhancement phases, brighter organs are marked with “+” symbols (more “+” indicates stronger enhancement), while darker regions—such as tumors—are marked with “–”.
    \textit{Non-contrast Phase:} before contrast injection, most organs have similar brightness.
    \textit{Arterial Phase:} after contrast injection, arteries, spleen, and pancreas brighten first.
    \textit{Venous Phase:} after the arterial phase, the liver, pancreas and spleen brightness appear stronger. Pancreatic tumors often remain darker in the arterial and venous phase, making them easier to detect.
    \textit{Delay Phase:} after the venous phase, kidneys reach peak enhancement, and liver enhancement gradually washes out.
    }
    \label{fig:supp_multi-phase}
\end{figure}

Blood-rich organs become bright in the arterial phase, most abdominal organs enhance in the venous phase, and contrast slowly fades in the delayed phase. Pancreatic tumors, however, usually stay darker than the surrounding pancreas in all these phases (see \figureautorefname~\ref{fig:supp_multi-phase}), creating a clear visual difference that does not appear in non-contrast scans. These predictable brightness changes make contrast phases crucial: they reveal tumors that would otherwise stay hidden and give radiologists multiple chances to confirm where the tumor is and how big it is.

\clearpage

\section{\method\ Implementations Details}
\label{sec: SMILE details}

\begin{figure}[ht]
    \centering
    \includegraphics[width=\linewidth]{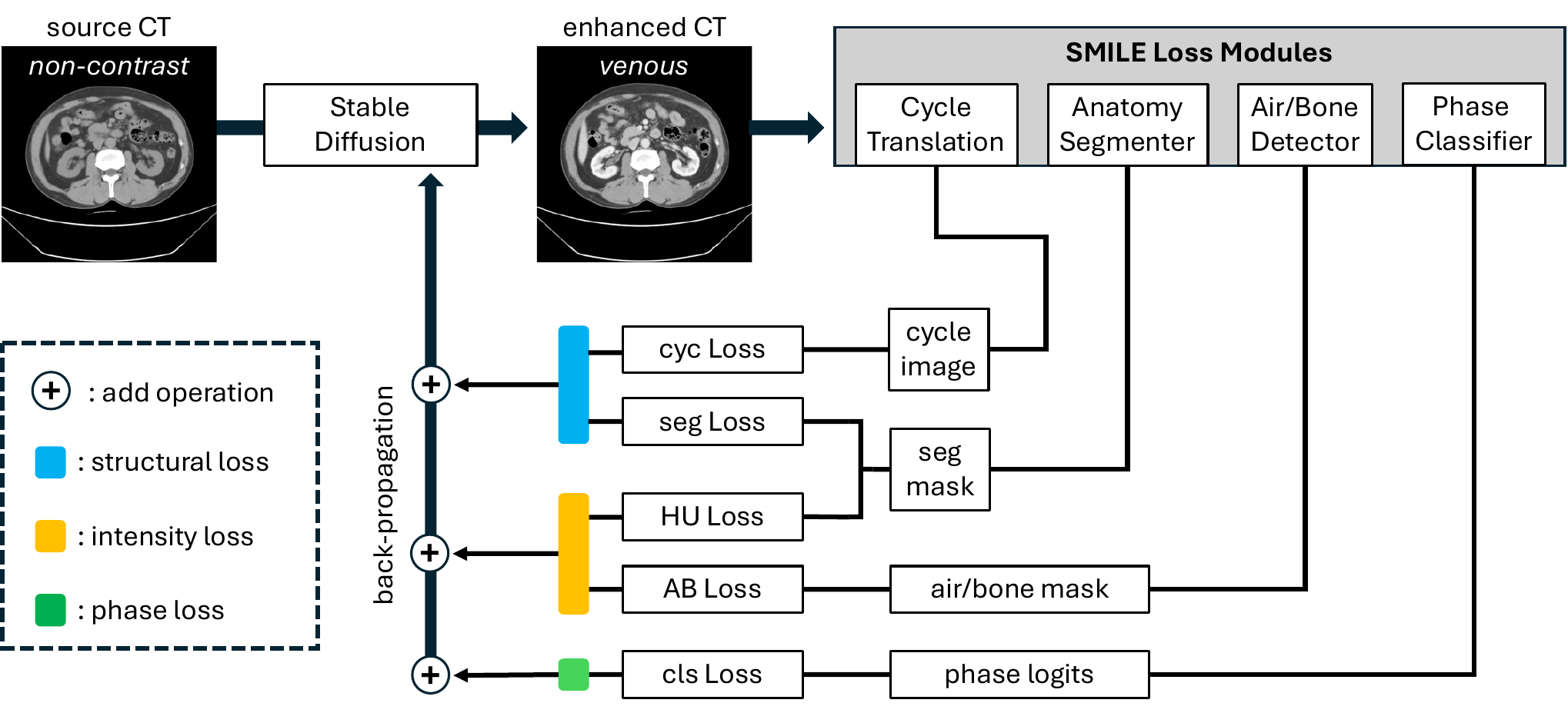}
    \caption{\textbf{Anatomical-aware design.} \method\ uses StableDiffusion~\cite{rombach2022high} as the diffusion backbone, and use unique anatomy-aware loss design to ensure enhancement quality. \emph{Structural loss:} Cycle translation and anatomy segmentation ensure anatomical structure preservation; \emph{Intensity loss:} HU and air/bone losses enforce realistic intensities; \emph{Phase loss:} A pretrained phase classifier ensures correct contrast phase.
    All losses back-propagate into the diffusion model to produce anatomically correct and phase-accurate enhanced CT scans.
}
    \label{fig:supp_method}
\end{figure}

\subsection{Why \method\ describes three loss fields but uses five loss weights?}
\method\ groups its supervision into three conceptual fields—structure, phase, and intensity—because these correspond to the three clinical requirements of contrast enhancement: preserving anatomy, achieving correct phase, and producing realistic tissue intensities.
However, As shown in \figureautorefname~\ref{fig:supp_method}, each field is implemented using practical loss components, resulting in five trainable losses: segmentation and cycle (structural), classifier (phase), and HU and air/bone (intensity).
Thus, to reflect these components precisely, the full objective is:

\begin{equation}
\mathcal{L}_{\text{SMILE}}
=
\mathcal{L}_{\text{diff}}
+ \lambda_{\text{seg}}\mathcal{L}_{\text{seg}}
+ \lambda_{\text{cyc}}\mathcal{L}_{\text{cyc}}
+ \lambda_{\text{cls}}\mathcal{L}_{\text{cls}}
+ \lambda_{\text{HU}}\mathcal{L}_{\text{HU}}
+ \lambda_{\text{AB}}\mathcal{L}_{\text{AB}}.
\end{equation}

\subsection{Why \method\ does not need registration?}
\method\ is trained on unregistered multi-phase CT by cleanly separating structural and intensity supervision. 
All geometry-related signals (segmentation + cycle consistency) come exclusively from the \textbf{source} CT, ensuring that no spatial information from the target phase is ever used. 
In contrast, the intensity and phase losses use only \textbf{alignment-free} cues from the target CT—mean organ HU values and slice-wise phase labels—which depend on global contrast behavior rather than voxel-wise correspondence. 
\method\ bypasses any need for voxel-level registration and remains robust to the natural misalignment between phases.

\subsection{Why \method\ introduces supervision progressively?}
We progressively activate loss modules during training to ensure stable optimization.
Early in training (0–2k steps), the model learns only the diffusion prior, preventing supervision signals from dominating before the generator produces meaningful structures. Phase and cycle losses are added at 10k steps to guide global enhancement behavior once the model forms basic anatomy. At 20k steps, segmentation, HU, and air/bone losses are enabled to refine structural boundaries and organ-level intensity patterns.

This staged strategy avoids unstable gradients and ensures that high-level supervision is added only when the model is ready to benefit from it, leading to more stable and reliable enhancement quality.Let $t$ denote the training step.  
The active loss at step $t$ is:
\begin{equation}\footnotesize
\mathcal{L}^{(t)} =
\begin{cases}
\mathcal{L}_{\text{diff}}, 
& 0 \le t < 2\text{k}, \\[6pt]
\mathcal{L}_{\text{diff}}
+ \lambda_{\text{cyc}}\mathcal{L}_{\text{cyc}}
+ \lambda_{\text{cls}}\mathcal{L}_{\text{cls}}, 
& 2\text{k} \le t < 20\text{k}, \\[6pt]
\mathcal{L}_{\text{diff}}
+ \lambda_{\text{cyc}}\mathcal{L}_{\text{cyc}}
+ \lambda_{\text{cls}}\mathcal{L}_{\text{cls}}
+ \lambda_{\text{seg}}\mathcal{L}_{\text{seg}}
+ \lambda_{\text{HU}}\mathcal{L}_{\text{HU}}
+ \lambda_{\text{AB}}\mathcal{L}_{\text{AB}},
& 20\text{k} \le t < 80\text{k}, \\[6pt]
\mathcal{L}_{\text{diff}}
+ \lambda_{\text{cyc}}^{(t)}\mathcal{L}_{\text{cyc}}
+ \lambda_{\text{cls}}^{(t)}\mathcal{L}_{\text{cls}}
+ \lambda_{\text{seg}}^{(t)}\mathcal{L}_{\text{seg}}
+ \lambda_{\text{HU}}^{(t)}\mathcal{L}_{\text{HU}}
+ \lambda_{\text{AB}}^{(t)}\mathcal{L}_{\text{AB}},
& t \ge 80\text{k},
\end{cases}
\end{equation}
where $\lambda_{i}^{(t)}$ are learnable weights estimated by the Uncertainty Loss Module.

\subsection{Why \method\ can adjust loss weights automatically?}
The idea of learnable loss weight is simple: if a supervision term is noisy or conflicting with others, its weight should be reduced; if a term is stable and provides consistent gradients, its weight should be increased.

Concretely, following standard uncertainty weighting for multi-task learning, we introduce one scalar parameter $s_i$ per loss term $\mathcal{L}_i \in \{\mathcal{L}_{\text{cyc}}, \mathcal{L}_{\text{cls}}, \mathcal{L}_{\text{seg}}, \mathcal{L}_{\text{HU}}, \mathcal{L}_{\text{AB}}\}$. The combined objective is written as
\begin{equation}
\mathcal{L}_{\text{SMILE}}
=
\mathcal{L}_{\text{diff}}
+ \sum_i \exp(-s_i)\,\mathcal{L}_i + s_i,
\end{equation}
and the effective weight for loss $\mathcal{L}_i$ at step $t$ is
$\lambda_i^{(t)} = \exp(-s_i)$.
During training, $s_i$ is optimized together with the network parameters via backpropagation. Intuitively, if a loss $\mathcal{L}_i$ is large or highly variable, the gradient w.r.t.\ $s_i$ will increase $s_i$ and thus decrease $\lambda_i^{(t)}$, down-weighting that term. Conversely, when a loss is small and consistent, $s_i$ is reduced, which increases its weight.

This design is reasonable for two reasons. \textbf{First,} it avoids manual tuning of five various loss scales (segmentation, cycle, classification, HU, and air/bone), which naturally live on different numeric ranges. \textbf{Second,} the weighting is data-driven: the model learns which supervisions are currently reliable instead of relying on a fixed heuristic schedule.

\clearpage

\section{\method\ Comparison in Early Tumor Detection with Baselines}
\begin{figure}[ht]
    \centering
    \includegraphics[width=\linewidth]{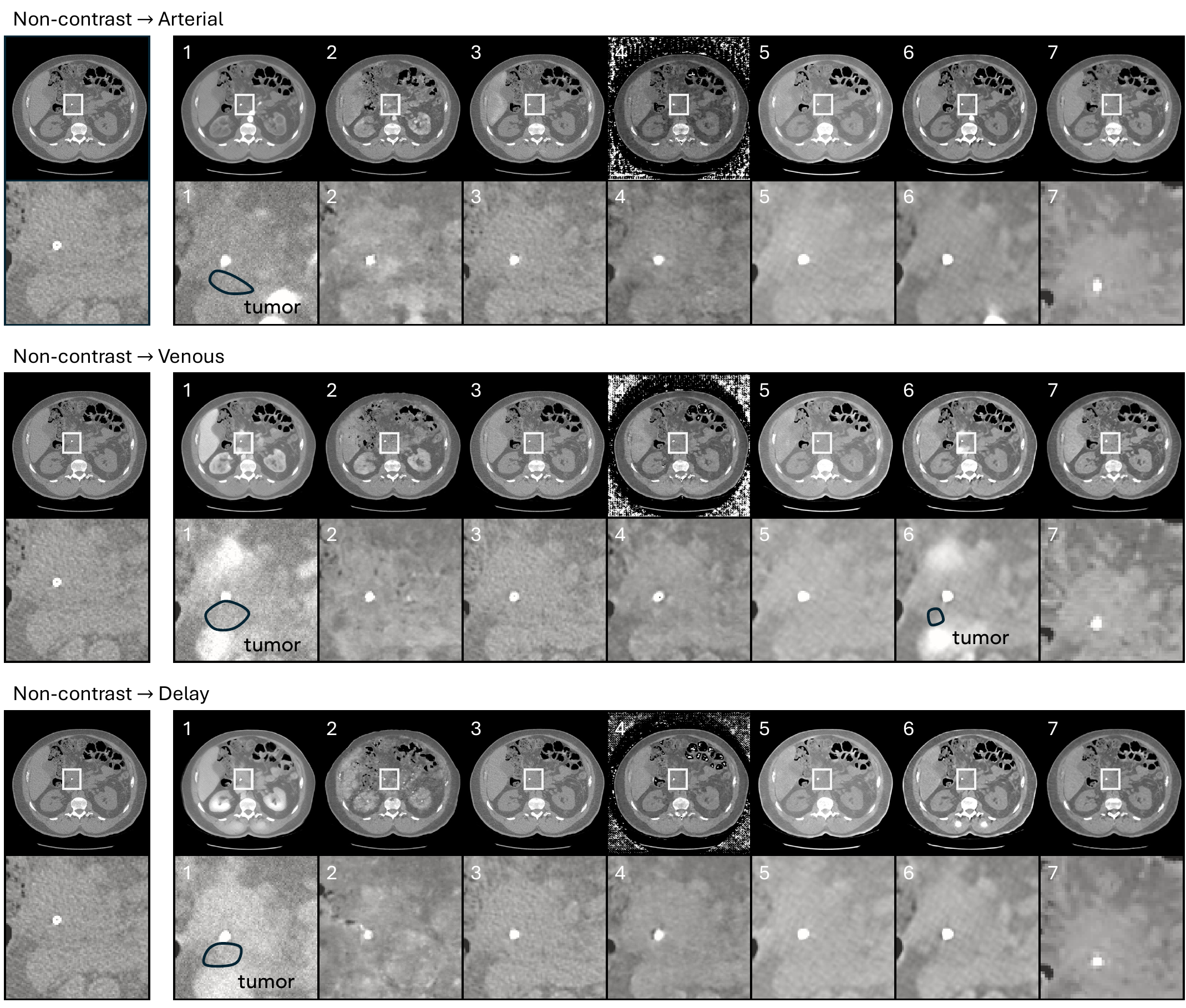}
    \caption{\textbf{\method\ outperforms all competing models in both anatomical integrity and tumor visibility (patient example \#1).} 
    For each source–target pair (non-contrast to arterial, venous, delay), the top row shows full slices and the bottom row zooms into the pancreas region.
    Methods are shown in the figure following this order:
    (1) \method, 
    (2) Pix2Pix~\cite{isola2017image}, 
    (3) CycleGAN~\cite{chu2017cyclegan}, 
    (4) CyTran~\cite{ristea2023cytran}, 
    (5) DALL-E~\cite{esser2021taming}, 
    (6) MedDiffusion~\cite{khader2023denoising}, 
    (7) CUT~\cite{park2020contrastive}. 
    }
    \label{fig: supp_vis_1}
\end{figure}

\begin{figure}[ht]
    \centering
    \includegraphics[width=\linewidth]{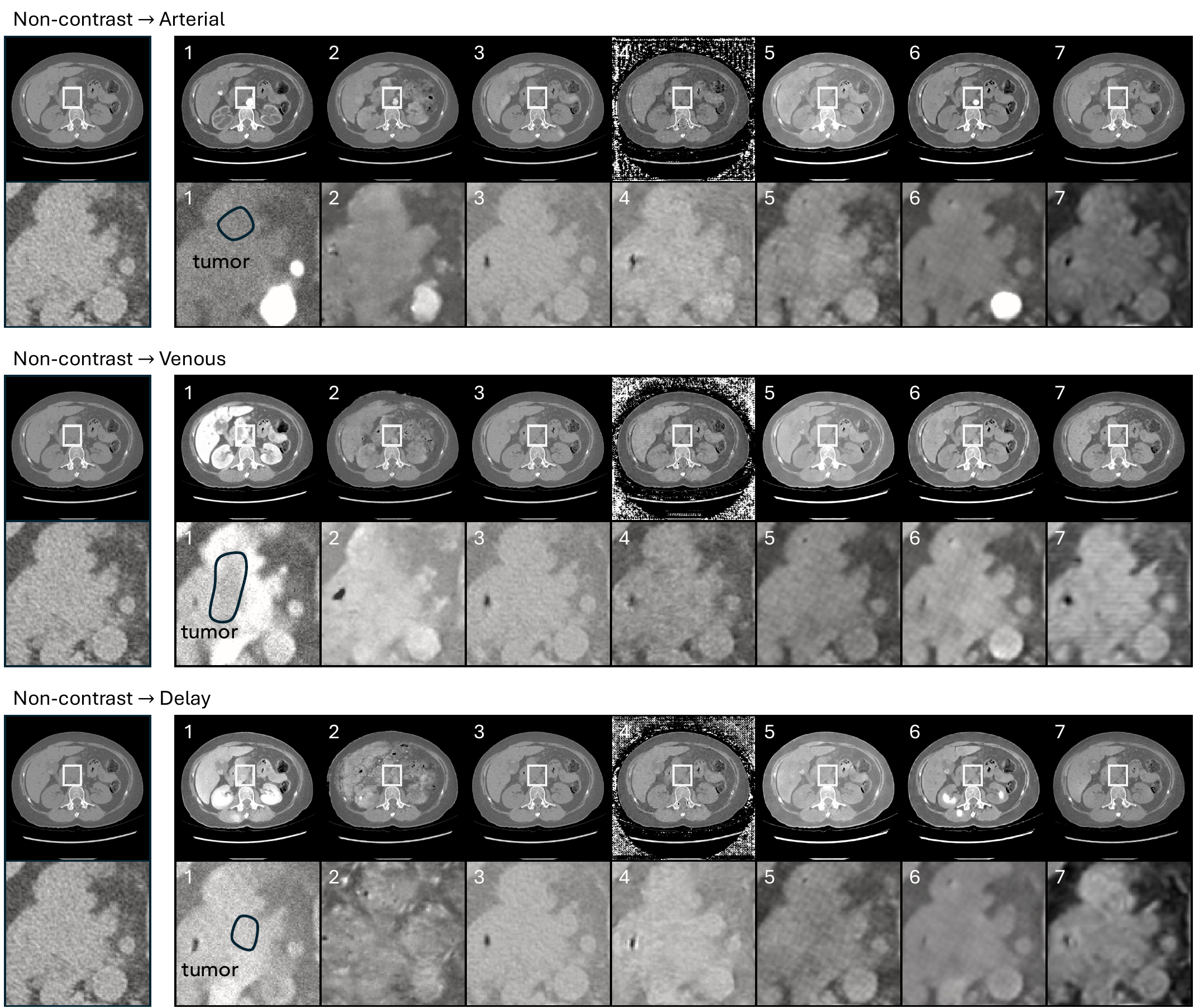}
    \caption{\textbf{\method\ outperforms all competing models in both anatomical integrity and tumor visibility (patient example \#2).} 
    For each source–target pair (non-contrast to arterial, venous, delay), the top row shows full slices and the bottom row zooms into the pancreas region.
    Methods are shown in the figure following this order:
    (1) \method, 
    (2) Pix2Pix~\cite{isola2017image}, 
    (3) CycleGAN~\cite{chu2017cyclegan}, 
    (4) CyTran~\cite{ristea2023cytran}, 
    (5) DALL-E~\cite{esser2021taming}, 
    (6) MedDiffusion~\cite{khader2023denoising}, 
    (7) CUT~\cite{park2020contrastive}. 
    \method\ improves tumor visibility, allowing the AI detector to identify the tumor reliably, while other baselines failed.}
    \label{fig:supp_vis_2}
\end{figure}


\clearpage
\section{\method\ Comparison in Enhancement Quality with Baselines (Ground-Truth Provided)}
\begin{figure}[ht]
    \centering
    \includegraphics[width=\linewidth]{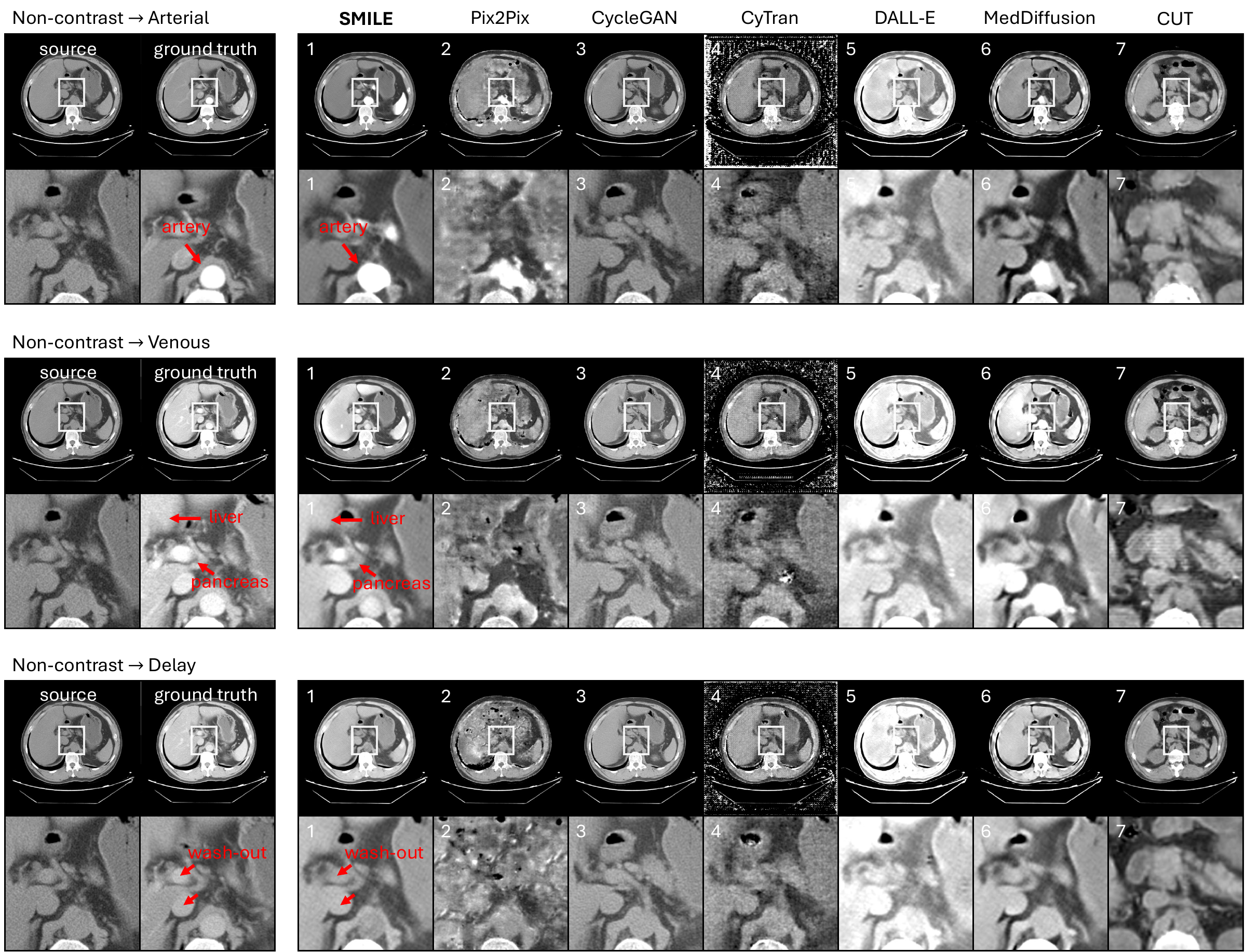}
    \caption{
    \textbf{\method\ produces the clearest and most phase-consistent enhancement (patient example \#3).}
    Above shows the qualitative comparison of non-contrast to arterial, venous, and delay enhancement, with \emph{registered ground truth provided}.
    The left-most column shows the source non-contrast CT and the ground-truth enhanced CT (arterial, venous, delay).
    All columns on the right show the enhanced results from different methods.
    For each source–target pair (non-contrast to arterial, venous, delay), the top row shows full slices and the bottom row zooms into the pancreas region.  
    We compare \method\ (label 1) with six representative baselines: 
    (2) Pix2Pix \cite{isola2017image}, 
    (3) CycleGAN \cite{chu2017cyclegan}, 
    (4) CyTran \cite{ristea2023cytran}, 
    (5) DALL-E \cite{esser2021taming}, 
    (6) MedDiffusion \cite{khader2023denoising}, 
    and (7) CUT \cite{park2020contrastive}.
    Baseline models often blur organ boundaries, introduce artifacts, or fail to reproduce the expected enhancement, especially around the pancreas and tumor region.
    On the contrary, across the three phase conversions—arterial, venous, and delay—\method\ shows the expected clinical patterns:
    In the arterial phase, SMILE correctly highlights the arteries, matching real contrast flow.
    In the venous phase, SMILE brightens the liver, pancreas, and veins, as seen in real venous enhancement.
    In the delay phase, SMILE reproduces the wash-out effect, where organ brightness slowly fades.
    These phase-specific changes match real CT behavior and help reveal tumors that are difficult to see in non-contrast scans. 
    }
    \label{fig:supp_vis_3}
\end{figure}

\begin{figure}[t]
    \centering
    \includegraphics[width=\linewidth]{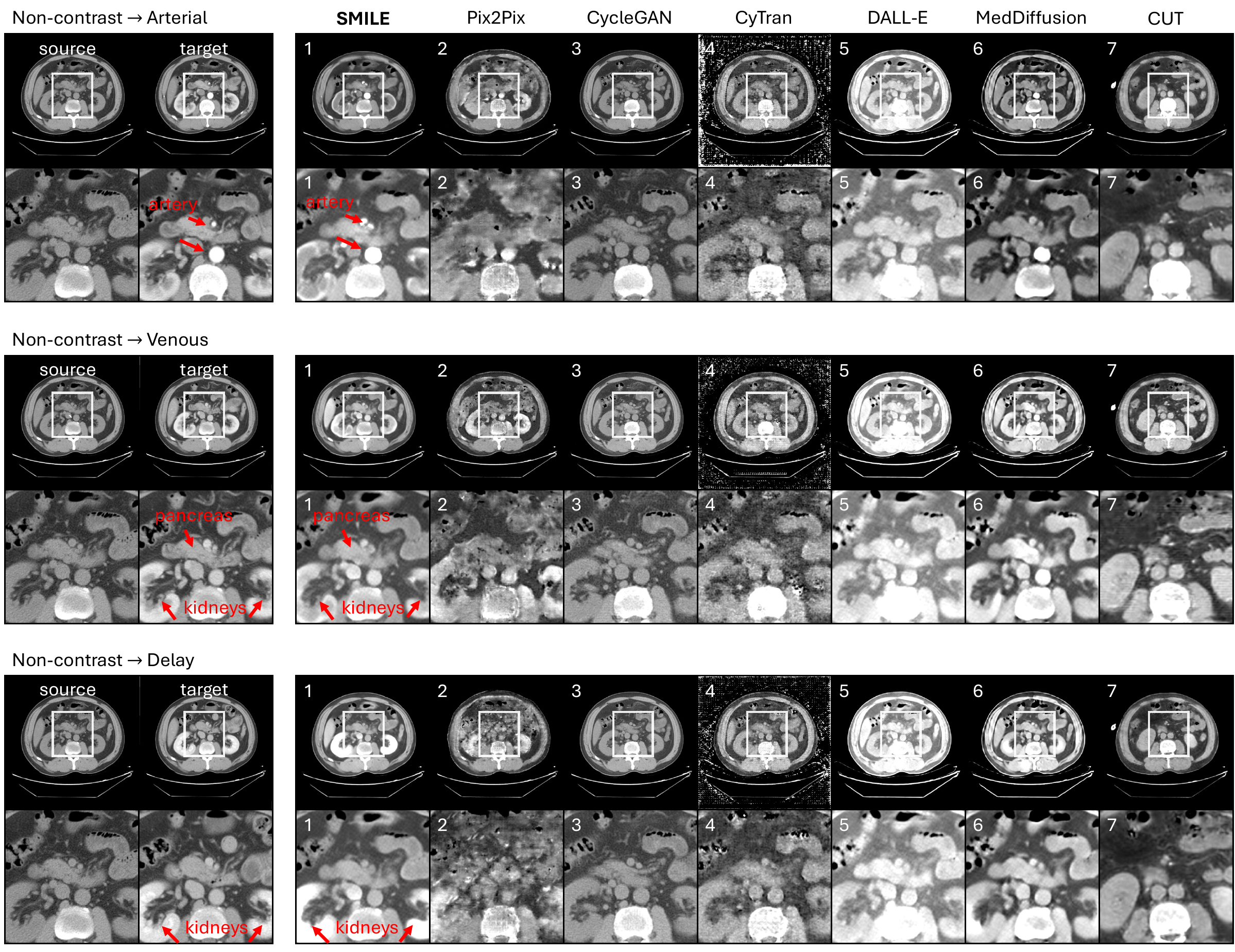}
    \caption{
    \textbf{\method\ produces the clearest and most phase-consistent enhancement (patient example \#4).}
    Above shows the qualitative comparison of non-contrast to arterial, venous, and delay enhancement, with \emph{registered ground truth provided}.
    The left-most column shows the source non-contrast CT and the ground-truth enhanced CT (arterial, venous, delay).
    All columns on the right show the enhanced results from different methods.
    For each source–target pair (non-contrast to arterial, venous, delay), the top row shows full slices and the bottom row zooms into the pancreas region.  
    We compare \method\ (label 1) with six representative baselines: 
    (2) Pix2Pix \cite{isola2017image}, 
    (3) CycleGAN \cite{chu2017cyclegan}, 
    (4) CyTran \cite{ristea2023cytran}, 
    (5) DALL-E \cite{esser2021taming}, 
    (6) MedDiffusion \cite{khader2023denoising}, 
    and (7) CUT \cite{park2020contrastive}.
    Baseline models often blur organ boundaries, introduce artifacts, or fail to reproduce the expected enhancement, especially around the pancreas and tumor region.
    On the contrary, across the three phase conversions—arterial, venous, and delay—\method\ shows the expected clinical patterns:
    In the arterial phase, SMILE correctly highlights the arteries, matching real contrast flow.
    In the venous phase, SMILE brightens the, pancreas, and veins, and part of kidney as seen in real venous enhancement.
    In the delay phase, SMILE correctly shows the wash-out pattern: only excretory organs such as the kidneys remain bright, while most other organs lose contrast.
    These phase-specific changes match real CT behavior and help reveal tumors that are difficult to see in non-contrast scans.
    }
    \label{fig:supp_vis_4}
\end{figure}

\clearpage
\section{\method\ Compared with Commercialized Generative Vision Models}
\begin{figure}[ht]
    \centering
    \includegraphics[width=\linewidth]{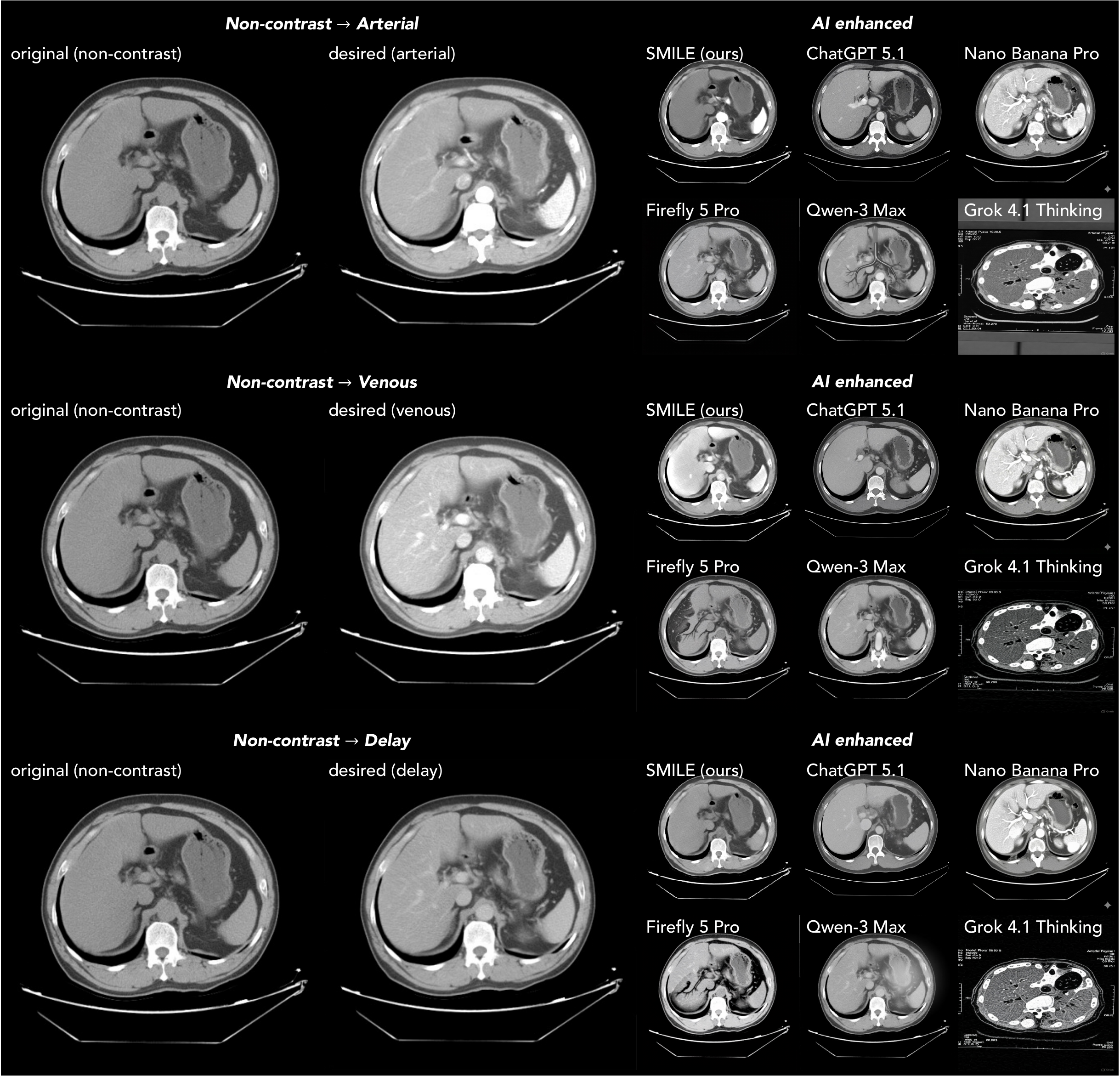}
    \caption{\method\ consistently produces the most anatomically accurate and clinically meaningful enhancement across all contrast phases, compares to current commercialized generative vision models.}
    \label{fig:supp_ai_1}
\end{figure}

\begin{figure}[ht]
    \centering
    \includegraphics[width=\linewidth]{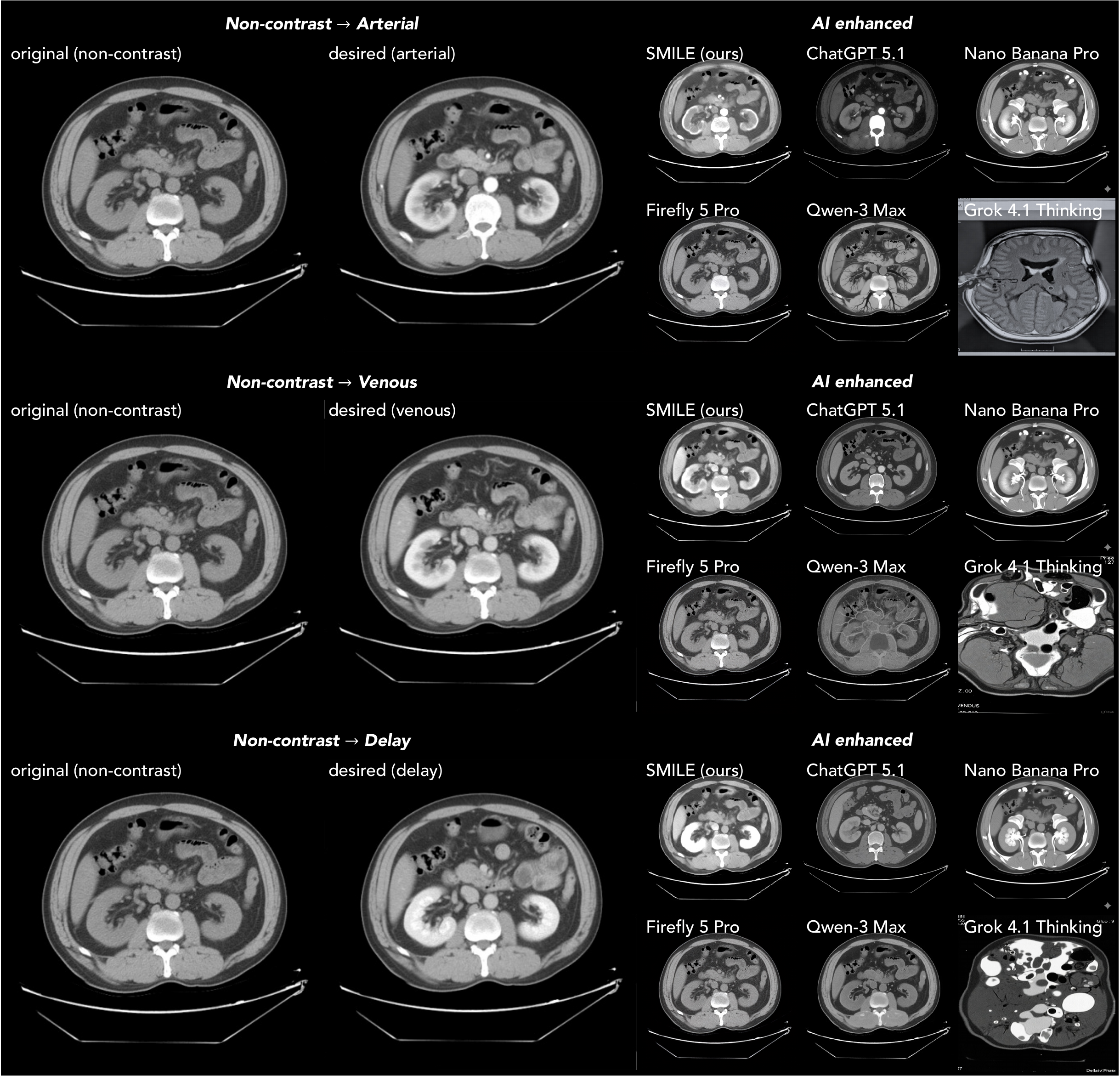}
    \caption{\method\ consistently produces the most anatomically accurate and clinically meaningful enhancement across all contrast phases, compares to current commercialized generative vision models.}
    \label{fig:supp_ai_2}
\end{figure}

\clearpage
\section{\method\ Performance on Specific Organs}
\begin{figure}[ht]
    \centering
    \includegraphics[width=\linewidth]{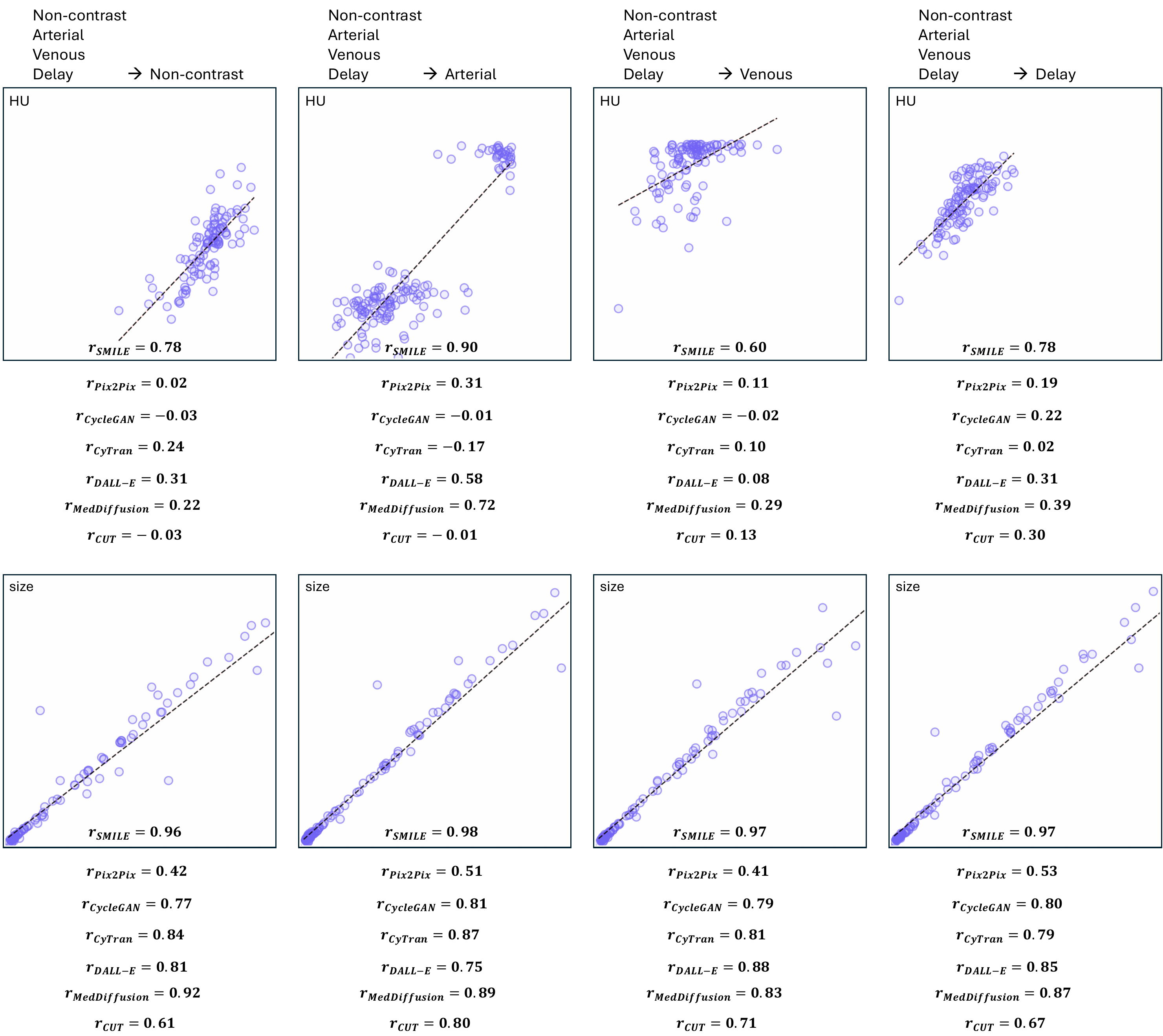}
    \caption{
    \textbf{Contrast enhancement of five key organs, i.e., aorta, liver, pancreas, kidneys, and spleen.}
    We evaluate \method’s enhancement accuracy by measuring HU and size correlation between enhanced CT and ground truth on selected organs. They are selected because these structures show the strongest contrast changes across phases and are clinically important for tumor diagnosis. Each column shows one enhancement target.
    \method\ achieves high HU and size correlations in all settings compared to other baselines. Performance is lower in the venous phase, which is expected because venous enhancement depends on subtle global perfusion patterns that are harder to learn. Even so, \method\ remains consistently better than all baseline methods. Overall, these results show that \method\ produces anatomically and intensity-consistent enhancement for the most clinically relevant organs.}
    \label{fig:placeholder}
\end{figure}

\end{document}